\documentclass[letterpaper]{article} 
\usepackage{aaai2026}  
\usepackage{times}  
\usepackage{helvet}  
\usepackage{courier}  
\usepackage[hyphens]{url}  
\usepackage{graphicx} 
\usepackage{natbib}  
\usepackage{caption} 
\usepackage{algorithm}
\usepackage{algorithmic}

\usepackage{newfloat}
\usepackage{listings}
\DeclareCaptionStyle{ruled}{labelfont=normalfont,labelsep=colon,strut=off} 
\floatstyle{ruled}
\newfloat{listing}{tb}{lst}{}
\floatname{listing}{Listing}
\makeatletter\def\@listi{\leftmargin\leftmargini \topsep .5em \parsep .5em \itemsep .5em}
\def\@listii{\leftmargin\leftmarginii \labelwidth\leftmarginii \advance\labelwidth-\labelsep \topsep .4em \parsep .4em \itemsep .4em}
\def\@listiii{\leftmargin\leftmarginiii \labelwidth\leftmarginiii \advance\labelwidth-\labelsep \topsep .4em \parsep .4em \itemsep .4em}\makeatother

\newcounter{checksubsection}
\newcounter{checkitem}[checksubsection]

\ifdefined\aaaianonymous
    \usepackage[submission]{aaai2026}  
\else
    \usepackage{aaai2026}              
\fi

\nocopyright

\usepackage{amsmath} 
\usepackage{amsfonts}
\usepackage{multirow} 
\usepackage{booktabs} 
\usepackage[cmyk,table]{xcolor}

\title{MoReMouse: Monocular Reconstruction of Laboratory Mouse}

\author{
    Yuan Zhong\equalcontrib\textsuperscript{\rm 1},
    Jingxiang Sun\equalcontrib\textsuperscript{\rm 1},\\
    Zhongbin Zhang\textsuperscript{\rm 1},
    Liang An\textsuperscript{\rm 1}\thanks{Corresponding authors},
    Yebin Liu\textsuperscript{\rm 1}\footnotemark[2]
}

\affiliations{
    \textsuperscript{\rm 1}Tsinghua University\\
    \{zhong-y19, zhangzb23\}@mails.tsinghua.edu.cn,
    \{anliang, liuyebin\}@mail.tsinghua.edu.cn,
    starkjxsun@gmail.com
}

\begin{document}

\maketitle

\begin{abstract}
Laboratory mice, particularly the C57BL/6 strain, are essential animal models in biomedical research. However, accurate 3D surface motion reconstruction of mice remains a significant challenge due to their complex non-rigid deformations, textureless fur-covered surfaces, and 
the lack of realistic 3D mesh models. 
Moreover, existing visual datasets for mice reconstruction only contain sparse viewpoints without 3D geometries.
To fill the gap, we introduce MoReMouse, the first monocular dense 3D reconstruction network specifically designed for C57BL/6 mice. To achieve high-fidelity 3D reconstructions, we present three key innovations. First, we create the first high-fidelity, dense-view synthetic dataset for C57BL/6 mice by rendering a realistic, anatomically accurate Gaussian mouse avatar. Second, MoReMouse leverages a transformer-based feedforward architecture combined with triplane representation, enabling high-quality 3D surface generation from a single image, optimized for the intricacies of small animal morphology. Third, we propose geodesic-based continuous correspondence embeddings on the mouse surface, which serve as strong semantic priors, improving surface consistency and reconstruction stability, especially in highly dynamic regions like limbs and tail.
Through extensive quantitative and qualitative evaluations, we demonstrate that MoReMouse significantly outperforms existing open-source methods in both accuracy and robustness.
\end{abstract}

\begin{links}
     \link{Project page}{https://zyyw-eric.github.io/MoreMouse-webpage/}
\end{links}

\begin{figure*}[t]
  \centering
  \includegraphics[width=\textwidth]{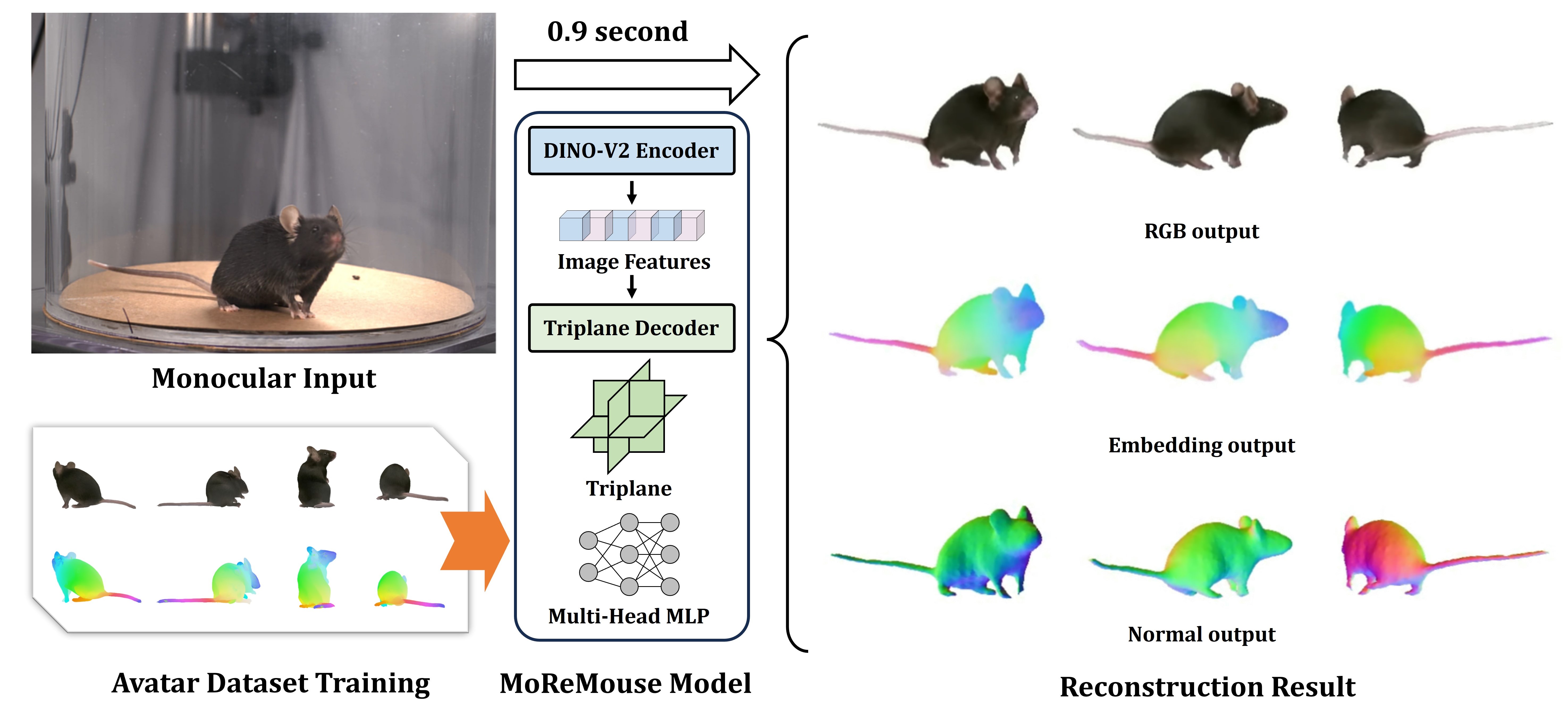}
  \caption{We present \textbf{MoReMouse}, the first monocular dense 3D reconstruction framework for laboratory mice. Given a single-view input (top-left, captured with an iPhone 15 Pro), MoReMouse predicts high-fidelity surface geometry and appearance within 0.9 seconds using a transformer-based triplane architecture (middle). 
  To assist model training, we render a dense-view synthetic dataset by building the first Gaussian mouse avatar from sparse-view real videos (bottom-left). Our method outputs RGB renderings, semantic embeddings, and normal maps from a single image (right). Corresponding video results are provided in the supplementary material.}
  \label{fig:teaser}
\end{figure*}


\section{Introduction}

Laboratory mice are widely used model organisms in biomedical research, especially for disease modeling, genetic studies, and neuroscience. Among them, the C57BL/6 strain is the most commonly used inbred line due to its well-characterized genome and relevance in neurodegenerative, oncological, and immunological research. Accurate 3D motion reconstruction of C57BL/6 mice is essential for behavioral phenotyping.

Over the last decade, deep learning has revolutionized both 2D and 3D animal pose estimation, with methods such as DeepLabCut~\cite{mathis2018deeplabcut}, SLEAP~\cite{pereira2022sleap} and DANNCE~\cite{dunn2021geometric} becoming standard tools across diverse species including mice. However, these methods are limited to sparse keypoint tracking and ignore important 3D surface motion. To this end, template-based papers have tried to build articulated mouse mesh models~\cite{49_bolanos2021three} and used them for mouse 3D tracking~\cite{an2023three}. However, the manually designed skinning weights are infeasible to represent the complex dynamic deformations of a mouse, resulting in severe self-penetration during its free movement. Moreover, mesh fitting methods require heavy and sensitive multi-view optimization, hindering its application for more convenient single view scenarios. Template-free animal reconstruction methods such as 3D-Fauna~\cite{li2024learning} lacks the knowledge of mouse-specific appearance and geometry, therefore has difficulty in recovering 3D C57BL/6 mice. 

The rapid evolution of 3D generative AI has blurred the boundary between 3D reconstruction and 3D generation, with large-scale datasets such as Objaverse~\cite{deitke2023objaverse} playing a crucial role in this trend. Fueled by such data, Large Reconstruction Models (LRMs)~\cite{hong2023lrm,tochilkin2024triposr,tang2024lgm} have demonstrated high efficiency and accuracy in single-view, feedforward 3D reconstruction of general objects. More recent generative frameworks, such as Trellis~\cite{xiang2025structured} based on rectified flow, and CLAY~\cite{zhang2024clay} based on diffusion, further push the scalability and controllability of 3D generation by learning powerful structured latent representations.
Although these models show great promise, they do not perform well on laboratory mice due to the lack of large-scale 3D mice supervision or dense-view mice images. 
Moreover, as these methods only considers single-image cases, no temporally semantic correspondences are ensured for long term dynamic object reconstruction. 


In this paper, we present \textbf{MoReMouse}, the first \textbf{mo}nocular dense 3D \textbf{re}construction model specifically designed for C57BL/6 laboratory \textbf{mouse}. MoReMouse is built upon transformer architectures and triplane representation, enabling efficient, feedforward 3D reconstruction, as shown in Fig.~\ref{fig:teaser}. To compensate for the scarcity of dense-view 3D mouse dataset, we construct the first animatable Gaussian avatar of mouse (\textbf{AGAM} for short) using an articulated mouse model~\cite{49_bolanos2021three} and Gaussian Splatting~\cite{kerbl2023gaussian}. 
Inspired by recent human avatar modeling~\cite{li2024animatable}, we use a 2D UV-based intermediate mouse pose representation to drive Gaussian point clouds. Thanks for the free-view rendering ability and diverse pose space of the avatar, we build a synthetic 64-view dataset, enabling effective training of MoReMouse network. 

Based on the new synthetic dataset, the framework of MoReMouse integrates multiple architectural and training enhancements inspired by LRMs. Specifically, we employ a two-stage training strategy, initially pretraining the model with triplane-NeRF~\cite{chan2022efficient,mildenhall2021nerf} representations, followed by fine-tuning with triplane-DMTet\cite{shen2021deep} for higher-fidelity reconstructions. To further improve the stability and robustness, we introduce geodesic-based feature embeddings, which enhance semantic consistency of surfaces across frames.

We evaluate MoReMouse on both synthetic and real-world datasets. The synthetic benchmark is rendered using our AGAM, while the real-world evaluation uses a 4-view video sequence of single freely moving C57BL/6 mouse captured by PointGray industrial cameras. 
Results demonstrate the superior performance of MoReMouse in monocular mouse reconstruction and novel view synthesis.
As a whole, MoReMouse represents a significant step toward practical and high-fidelity monocular 3D reconstruction of C57BL/6 mice, with potential to support broader research in computational modeling and analysis of small animals.
To summarize, our key contributions include:
\begin{itemize}
    \item We introduce \textbf{MoReMouse}, the first monocular dense 3D reconstruction model tailored for laboratory mice, specifically the C57BL/6 strain, and show its superiority over existing open-source alternatives.
    \item We develop the first animatable Gaussian avatar model for mice and use it to generate a large-scale, dense-view synthetic dataset for training and evaluation.
    \item We integrate geodesic-based continuous correspondence embeddings into feedforward reconstruction, improving single-view surface stability and detail preservation.
\end{itemize}

\section{Related work}
\subsection{3D Representation of Mouse}
Previously, most methods reconstruct mouse 3D keypoints by algebraic triangulation~\cite{35_nath2019using,36_karashchuk2021anipose,han2024multi,monsees2022estimation}, voxel-based triangulation~\cite{dunn2021geometric} or even by optical markers~\cite{marshall2021pair}. However, 3D keypoints provide only sparse geometric structures, neglecting essential dense surface information.
To represent the mouse surface, ~\citep{49_bolanos2021three} proposes an articulated mesh model based on computed tomography scans, and is used for multi-view mesh fitting~\cite{an2023three}. 
ArMo~\cite{bohnslav2023armo} presents a similar mouse mesh model, but lacks true anatomical bones. For rats, which are similar to mice in shape, similar mesh models are built~\cite{aldarondo2024virtual,klibaite2025mapping}. All these mesh models are limited to the manually designed skinning weights, which could not capture the realistic dynamic surface motion of mice.

\subsection{Monocular Reconstruction of Animals}
Since the proposal of SMAL model which represents several quadrupedal species~\cite{63_zuffi20173d}, predicting the pose and shape parameters of SMAL from images has dominated the monocular reconstruction of animals~\cite{zuffi2019three,biggs2020wldo,xu2023animal3d,rueegg2022barc,lyu2025animer}. Although they perform well for quadrupeds, they cannot reconstruct rodents due to the limited shape space of SMAL. The AWOL model~\cite{zuffi2024awol} extends SMAL to represent rodents using CLIP~\cite{radford2021learning}, yet its pose and shape realism is still limited. 
Some recent papers try to build animal avatars from videos by augmenting the skinning weights or adding additional deformations, such as BANMo~\cite{yang2022banmo}, Gart~\cite{lei2024gart} or AnimalAvatars~\cite{Sabathier2024AnimalAR}.  
However, these methods all require a heavy optimization process during inference, and rely on dense pose estimation or accurate SMAL parameters for initialization.
Template-free methods such as MagicPony~\cite{wu2023magicpony} and 3D Fauna~\cite{li2024learning} leverage collections of single-view images of deformable species captured in the wild to jointly learn canonical models and pose regressors. 
However, these approaches do not generalize well to mice due to domain and geometry gaps. 

\subsection{Large Reconstruction Model}
Large reconstruction models have recently achieved substantial progress in generating high-quality 3D content~\cite{hong2023lrm,Li2023Instant3DFT,zou2024triplane,gslrm2024,sun2024dreamcraft3d++}. These models are typically trained in an end-to-end fashion to predict implicit 3D representations, often through triplane-based neural radiance fields. 
Recent methods improve quality and efficiency by learning generative models over compact latent codes derived via VAE-like encoders~\cite{hu2025simulating}. Some focus purely on shape~\cite{zhang20233dshape2vecset,li2024craftsman3d}, while others incorporate appearance~\cite{gupta20233dgen,xiong2024octfusion}. CLAY~\cite{zhang2024clay} presents a unified system for geometry and PBR texture generation. Trellis~\cite{xiang2025structured} introduces structured latent codes via rectified flow transformers, jointly modeling geometry and appearance for versatile decoding into NeRFs, Gaussians, and meshes.
However, these methods generally depend on access to 3D supervision~\cite{deitke2023objaverse,wu2023omniobject3d}, such as ground-truth shapes or high-quality multi-view data, during training. The data scarcity is one of the main challenges for training an LRM for laboratory mice. 

\section{Method}
Our goal is to reconstruct dense surface geometry of a laboratory mouse from a single image in a feedforward manner. We focus specifically on the C57BL/6 strain, a widely used model organism in biomedical research.
We first introduce our Gaussian mouse avatar (AGAM) and synthetic dataset in Sec.~\ref{sec:sec:dataset}, then the structure of MoReMouse in Sec.~\ref{sec:sec:moremouse}. 

\subsection{Synthetic Dataset using Gaussian Avatar}
\label{sec:sec:dataset}
\begin{figure}[t]
  \centering
  \includegraphics[width=0.8\linewidth]{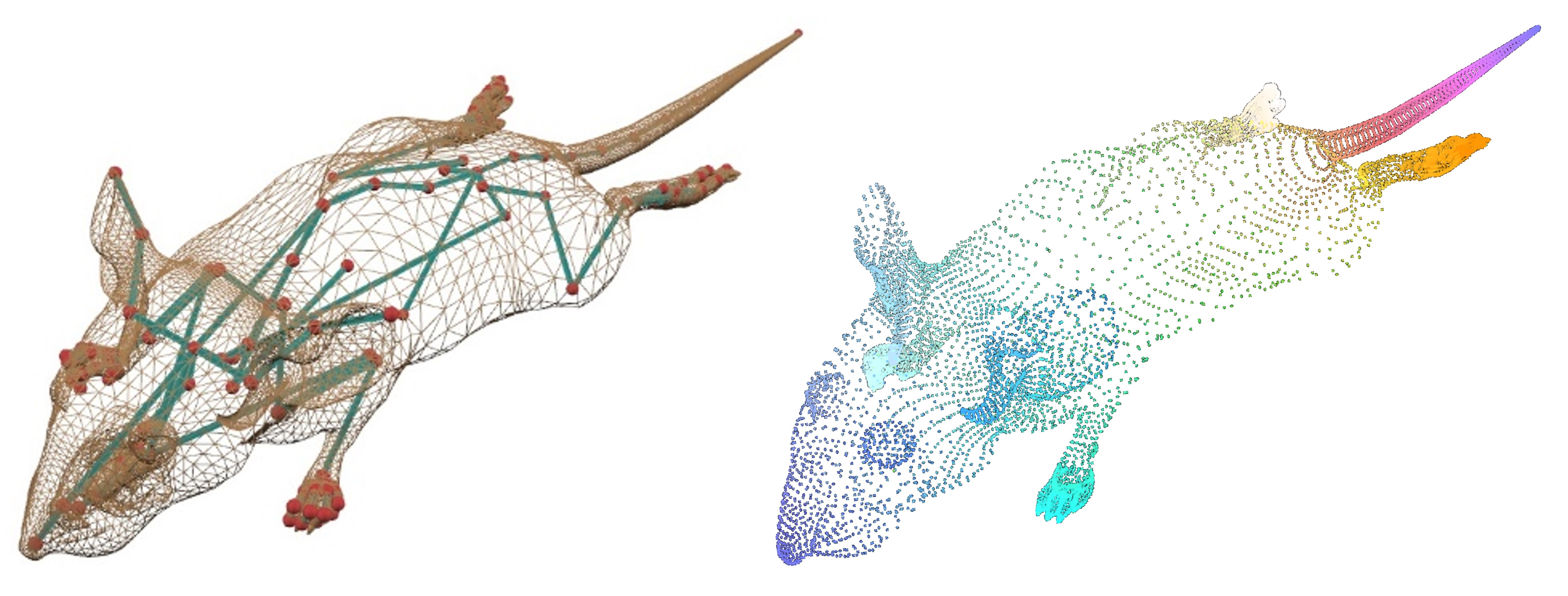}
  \caption{Left: the articulated mouse mesh we used. Right: color-coded geodesic feature embedding, promoting surface separability and reconstruction consistency.}
  \label{fig:mouse_mesh_model}
\end{figure}

\subsubsection{Mouse Parametric Model} 
\label{sec:sec:sec:mouse_articulation}

As shown in Fig~\ref{fig:mouse_mesh_model}, we modify the mouse model from ~\citep{49_bolanos2021three} following the approach of ~\citep{an2023three}, where redundant mesh vertices (teeth and tone) are removed. The model consists of \(N_J = 140\) articulated joints and \(N_V = 13059\) vertices, and conforms to Linear Blend Skinning (LBS) for posing. Poses of this refined model serve as the avatar control signals. The whole parameter set is denoted as \( \Psi \), including joint rotations \( \theta \in \mathbb{R}^{3N_J} \), global translation \( t \in \mathbb{R}^{3} \), global rotation \( r \in \mathbb{R}^{3} \), global scale \( s \in \mathbb{R} \) and bone length deformation parameters \( B \in \mathbb{R}^{21} \) (Please refer to ~\cite{an2023three} for more details). We decompose \( \Psi \) into local parameters \( \Psi_l = \{\theta, B\} \) and global parameters \( \Psi_g = \{r, t, s\} \) for the sake of simplicity.

\subsubsection{Gaussian Mouse Avatar}
\label{sec:sec:sec:avatar}

\begin{figure}[t]
  \centering
  \includegraphics[width=1.0\linewidth]{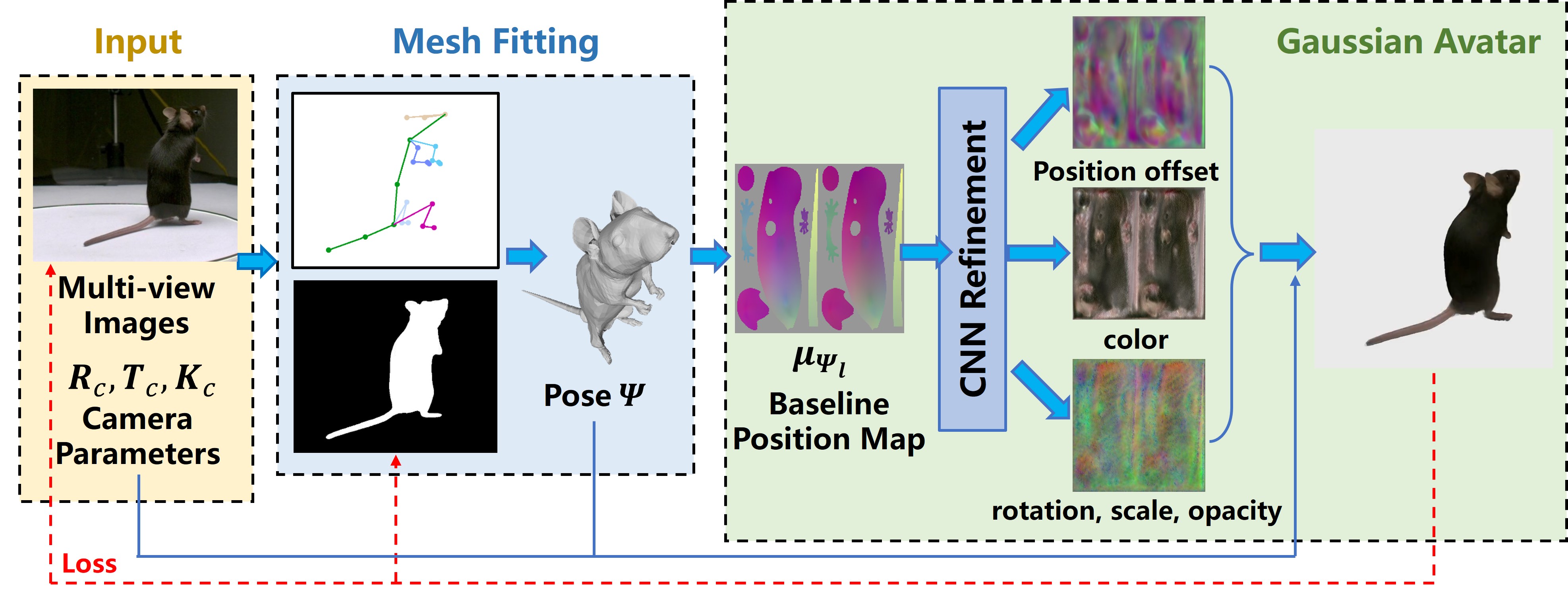}
  \caption{Pipeline of the Gaussian Mouse Avatar Generation. The process begins with Mesh Fitting, followed by Gaussian Avatar training. ``CNN'' here is StyleUNet. 
  }
  \label{fig:mouse_avatar_pipeline}
\end{figure}

Gaussian point clouds~\cite{kerbl2023gaussian} use an explicit parameterization for 3D representation. Each Gaussian point is characterized by a mean position \( \mu \in \mathbb{R}^3 \) and an anisotropic covariance matrix \( \Sigma \): 
\begin{equation} 
    G(x) = e^{-\frac{1}{2} (x - \mu)^T \Sigma^{-1} (x - \mu)}. 
\end{equation}

To facilitate optimization, the covariance matrix is reformulated as $   \Sigma = R S S^T R^T$, where \( S \) is a diagonal scaling matrix controlled by a three-dimensional vector \( s \in \mathbb{R}^3 \), and the rotation matrix \( R \) is represented by a normalized quaternion \( q \in \mathbb{R}^4 \). Additionally, each Gaussian encodes a color \( c \in \mathbb{R}^3 \) and an opacity \( o \in \mathbb{R} \). During rendering, Gaussians within the pixel projection region are sorted by depth and composited using alpha blending.

The pipeline of our AGAM is shown in Fig.~\ref{fig:mouse_avatar_pipeline}. 
Our key idea is to parameterize the mouse UV texture, mapping each valid texel in the UV space to a corresponding Gaussian point, where the texture color represents the Gaussian’s position. 
To achieve global-pose invariant controlling of AGAM, 
we set \( \Psi_g = 0 \) and drive the mesh model using only \( \Psi_l \), resulting in canonical vertex coordinates \( V_{\Psi_l} \). The UV-mapped positions of these vertices are used to interpolate Gaussian locations via barycentric coordinates. This defines the baseline Gaussian position map \( \mu_{\Psi_l} \):
\begin{equation} 
    \mu_{\Psi_l} = \mathcal{D}_{\Psi_l} (\mu_0),
\end{equation}
where \( \mu_0 \) is the UV coordinate map in the neutral pose, and \( \mathcal{D}_{\Psi_l} \) represents the LBS deformation mapping induced by the skeletal model.

The core network of AGAM is a set of StyleUNets~\cite{wang2023styleavatar} that predict the position offset map \( \Delta \mu_{\Psi_l} \) and other Gaussian attributes \( c_{\Psi_l} \), \( o_{\Psi_l} \), \( q_{\Psi_l} \), and \( s_{\Psi_l} \) conditioned on the UV coordinate map \(\mu_{\Psi_l}\). Note that, the Gaussian positions $\mu'_{\Psi_l}$ are derived by $\mu'_{\Psi_l} = \mathcal{D}_{\Psi_l} (\mu_0 + \Delta \mu_{\Psi_l}).$
Then the Gaussian Splatting rendering function generates the view-dependent rendered image \( I_c^{\text{rend}} \):
\begin{equation} 
    I_c^{\text{rend}} = GS(R_c, T_c, K_c, \mu'_{\Psi_l}, c_{\Psi_l}, q'_{\Psi_l}, s_{\Psi_l}),
\end{equation}
where \( GS(\cdot) \) denotes the Gaussian splatting renderer, and \( R_c, T_c, K_c \) are camera parameters.

\subsubsection{Avatar Training}
\label{sec:sec:sec:avatar_training}
We define loss functions between the real image \( I_c \) and the rendered image \( I_c^{\text{rend}} \) as:
\begin{equation} 
\begin{aligned}
\mathcal{L} &= \mathcal{L}_1(I_c - I_c^{\text{rend}}) + \lambda_{\text{SSIM}} (1 - \mathcal{L}_{\text{SSIM}}(I_c, I_c^{\text{rend}})) \\
&\quad + \lambda_{\text{LPIPS}} \mathcal{L}_{\text{LPIPS}}(I_c, I_c^{\text{rend}}),
\end{aligned}
\end{equation}
where \( \mathcal{L}_{\text{SSIM}} \) is structural similarity loss, \( \mathcal{L}_{\text{LPIPS}} \) is perceptual similarity loss, \( \lambda_{\text{SSIM}} = 0.2 \) and \( \lambda_{\text{LPIPS}} = 0.1 \). We use Total Variation (TV) loss to enforce point smoothness. See the supplementary material for further details.

Since C57BL/6 mice usually show similar texture, 
we only use a 6-view mouse video sequence ``markerless\_mouse\_1'' proposed by \cite{dunn2021geometric} for avatar training. This sequence captures 18,000 frames of a single freely moving mouse in an open field at 1152x1024@100FPS, where the mouse mesh models are borrowed from ~\cite{an2023three}. Note that, we only use 800 frames for training, which are uniformly sampled from the first 8000 frames of ``markerless\_mouse\_1". Avatar training takes 400k steps. 

\subsubsection{Dense-view Dataset}
We drive AGAM using the poses from all frames of the ``markerless\_mouse\_1'' video sequence, and render evenly distributed 64-view synthetic images. Among the full 18,000 frames, we use the first 6,000 frames for training and the last 6,000 frames for testing. Note that, the 800 frames used for avatar creation all belong to the first 6000 frames, therefore the poses in testset are unseen. To construct the training set, we sample two sets of 64 viewpoints per frame, rendering 12,000 multi-view scenes in total. The optical axes of these viewpoints point at the scene origin. Please refer to the supplementary file for more information about view selection and data samples.

\subsection{MoReMouse Architecture}
\label{sec:sec:moremouse}
\subsubsection{Transformer and Triplane Representation}
\label{sec:sec:sec:moremouse_network}
\begin{figure}[ht]
  \centering
  \includegraphics[width=1.0\linewidth]{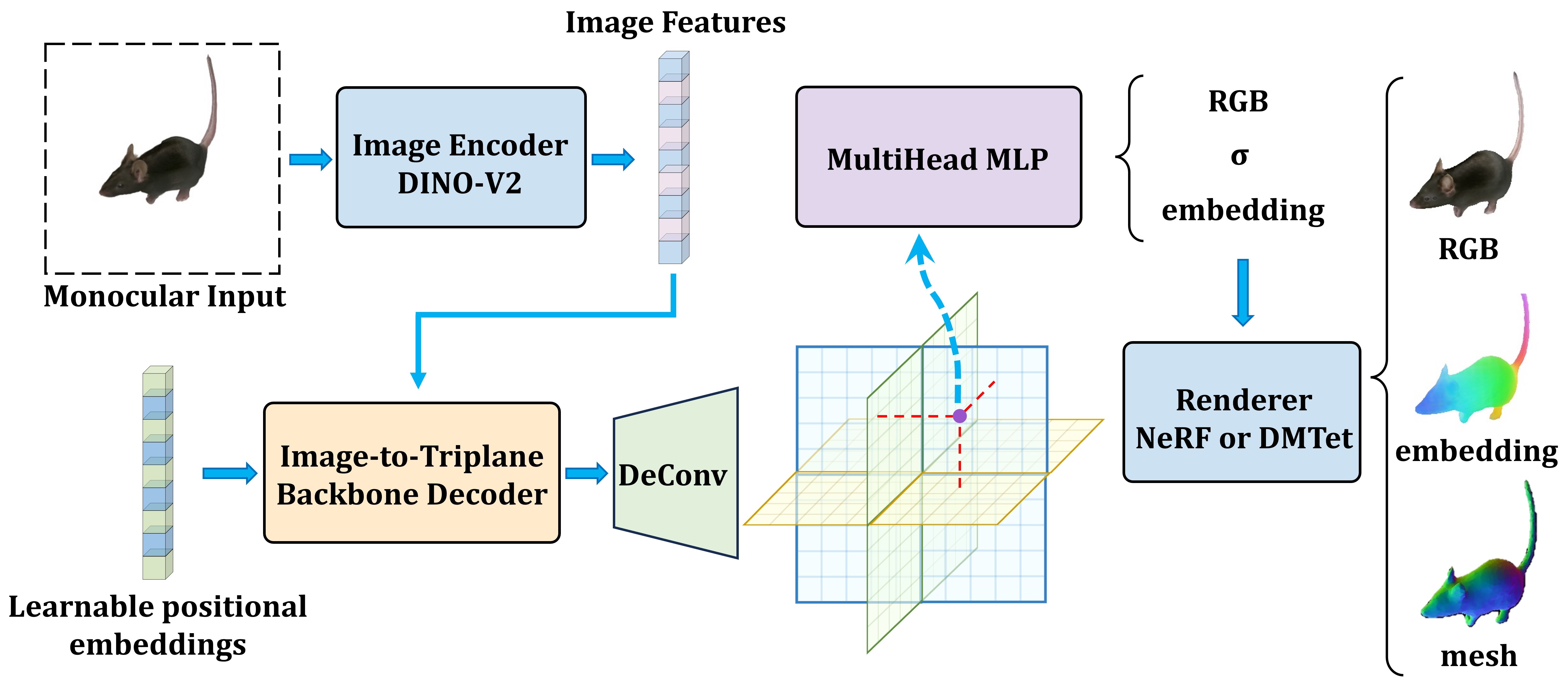}
    \caption{Overview of the MoReMouse architecture. Given a single image, a DINOv2 encoder and transformer decoder generate triplane features, which are queried to produce color, density, and embeddings. Rendering is performed via NeRF or DMTet for volumetric or surface outputs.}
   \label{fig:moremouse_architecture}
\end{figure}

Similar to LRM~\cite{hong2023lrm}, MoReMouse adopts a Transformer and triplane-based architecture, designed for monocular 3D reconstruction. As illustrated in Fig.~\ref{fig:moremouse_architecture}, the model consists of three key components: an image encoder, an image-to-triplane decoder, and a triplane-based geometric representation that integrates both NeRF~\cite{mildenhall2021nerf} and DMTet~\cite{shen2021deep} for implicit and explicit geometry modeling.

We use a pretrained DINOv2 encoder to extract latent features from the input image. These features are decoded via a transformer with cross-attention to generate a triplane representation, which encodes color, density, and semantic attributes for any 3D query point.



During the initial training stage, we employ NeRF for volumetric rendering without introducing explicit geometry, enabling smooth gradient propagation through the entire density field. After convergence, we switch to DMTet to obtain an explicit surface representation for fine-level geometric supervision.

The fine-scale structures of mice, particularly in the limbs and tail, present challenges for high-fidelity reconstruction in a feed-forward manner. To address this, we modify the LRM framework by increasing the triplane resolution to 64 to improve detail preservation. Additional architectural details are provided in the supplementary materials.

\subsubsection{Feature Embedding for Surface Consistency}
\label{sec:sec:sec:embedding}

Mouse reconstruction is challenging due to black fur with weak texture cues and fine-scale surface details. Inspired by Animatable Gaussians~\cite{li2024animatable}, we incorporate feature embeddings computed from geodesic distances~\cite{shamai2017geodesic} into the surface representation. These embeddings are used as additional color channels during training, serving as semantic priors to enhance reconstruction stability.

As shown in Fig.~\ref{fig:mouse_mesh_model}, given the set of mouse model vertices \( V \in \mathbb{R}^{N_V \times 3} \), we compute the pairwise geodesic distances between vertices to obtain the geodesic distance matrix:
\begin{equation} 
    G_V \in \mathbb{R}^{N_V \times N_V}, \quad G_V(i, j) = d_{\text{geo}}(V_i, V_j),
\end{equation}
where $d_{\text{geo}}(V_i, V_j)$ represents the geodesic distance between vertices $V_i$ and $V_j$.

To learn a continuous correspondence embedding that captures the intrinsic structure of the mouse surface, we define the embedding matrix \( E \in \mathbb{R}^{N_V \times 3}\)
and optimize it by minimizing the $L_2$ loss between the embedding-derived distance matrix $G_E$ and the geodesic distance matrix $G_V$:
\begin{equation} 
    \mathcal{L}_{\text{geo}} = \sum_{i,j} \left\| G_E(i, j) - G_V(i, j) \right\|_2^2,
\end{equation}
where $G_E(i, j) = \|E_i - E_j\|_2$ represents the Euclidean distance in the embedding space.

However, directly using these embeddings as color channels is suboptimal. The three-dimensional embeddings often form an elongated mouse-like shape in feature space, failing to fully utilize the available color space. Additionally, large dark or bright regions appear, lacking sufficient contrast against the black or white backgrounds used during rendering. To address this, we perform Principal Component Analysis (PCA) on the embeddings, selecting the first two principal components as the hue (H) and saturation (S) values in the HSV color space, while keeping the value (V) channel fixed at 1:
\begin{equation} 
    H = \text{PCA}_1(E), \quad S = \text{PCA}_2(E), \quad V = 1.
\end{equation}
This transformation improves the separability of the embeddings in the rendered images.


\subsubsection{Training and Optimization}
\label{sec:sec:sec:moremouse_train}

During training, two randomly selected viewpoints from each scene serve as the input-output pair. To mitigate spatial ambiguity, we fix the mouse position at the scene origin. In preprocessing, we also apply translation and scaling transformations to the input images, ensuring that the mouse centroid is aligned with the image center and normalized to a consistent scale. Experimental results show that MoReMouse generalizes well from training on fixed perspectives to varying viewpoints in real-world scenarios.

We employ a two-stage training strategy, where NeRF-based volumetric rendering is first trained for 60 epochs to ensure proper density field convergence, followed by DMTet-based explicit geometry modeling for an additional 100 epochs to refine surface details. Details of the training loss terms and hyperparameters are provided in the supplementary material.

\section{Experiments}
\subsection{Experimental Setup}

We evaluate MoReMouse on two types of testing datasets: a synthetic benchmark and a real-world captured dataset.

\paragraph{Synthetic Evaluation Data.}
We use the last 6,000 frames from the ``markerless\_mouse\_1'' sequence to generate a synthetic multi-view dataset consisting of 4 orthogonal viewpoints. These views are rendered using our Gaussian avatar pipeline and serve as a controlled evaluation set where ground-truth camera parameters and rendering conditions are known.

\paragraph{Real Captured Data.}
To assess performance under realistic conditions, we collect a real-world multi-view dataset using PointGray industrial cameras. This dataset contains 5,400 frames of freely moving mice captured from four calibrated viewpoints. More details about camera setup and calibration are provided in the supplementary materials.

\paragraph{Baselines.}
Due to the lack of monocular 3D reconstruction methods specifically designed for mice, we compare MoReMouse against four general-purpose single-view reconstruction baselines:

\begin{itemize}
    \item \textbf{TripoSR}~\cite{tochilkin2024triposr}: A state-of-the-art open-source Large Reconstruction Model (LRM) known for strong single-view 3D generation performance.
    \item \textbf{Triplane-GS}~\cite{zou2024triplane}: A LRM variant based on Gaussian representations.
    \item \textbf{LGM}~\cite{tang2024lgm}: A multi-view diffusion-based method that reconstructs 3D Gaussian point clouds.
    \item \textbf{InstantMesh}~\cite{xu2024instantmesh}: A mesh-based 3D reconstruction method using multi-view diffusion priors.
\end{itemize}


All methods are evaluated on the task of novel view synthesis, since obtaining ground-truth 3D geometry for mouse subjects is impractical. Additionally, we include surface geometry metrics (IoU against multi-view visual hulls) in the supplementary material. We use Track-Anything~\cite{yang2023trackanythingsegmentmeets} to remove background regions in real videos.


\begin{table}[h]
\centering
\renewcommand{\arraystretch}{1.2}
\setlength{\tabcolsep}{4pt}
\begin{tabular}{llccc}
\toprule
\textbf{Dataset} & \textbf{Method} & \textbf{PSNR} $\uparrow$ & \textbf{SSIM} $\uparrow$ & \textbf{LPIPS} $\downarrow$ \\
\midrule
\multirow{5}{*}{\textit{synthetic}} 
& \textbf{Ours} & \cellcolor{cyan!25}22.027 & \cellcolor{cyan!25}0.9660 & \cellcolor{cyan!25}0.05279 \\
& TripoSR & 13.673 & 0.8032 & 0.18255 \\
& Triplane-GS & \cellcolor{cyan!10}18.049 & \cellcolor{cyan!10}0.9268 & \cellcolor{cyan!10}0.10151 \\
& LGM & 14.460 & 0.8805 & 0.13274 \\
& InstantMesh & \cellcolor{cyan!5}15.821 & \cellcolor{cyan!5}0.8987 & \cellcolor{cyan!5}0.11312 \\
\midrule
\multirow{5}{*}{\textit{real}} 
& \textbf{Ours} & \cellcolor{cyan!25}18.422 & \cellcolor{cyan!25}0.9478 & \cellcolor{cyan!25}0.08674 \\
& TripoSR & 11.518 & 0.8114 & 0.19672 \\
& Triplane-GS & \cellcolor{cyan!10}16.789 & \cellcolor{cyan!10}0.9298 & \cellcolor{cyan!10}0.11002 \\
& LGM & \cellcolor{cyan!5}15.215 & \cellcolor{cyan!5}0.9197 & 0.12838 \\
& InstantMesh & 15.631 & \cellcolor{cyan!5}0.9175 & \cellcolor{cyan!10}0.11339 \\
\bottomrule
\end{tabular}
\caption{Quantitative results on both synthetic and real-world datasets. 
MoReMouse achieves the best performance across all metrics in both domains.} 
\label{tab:quant_combined}
\end{table}

\begin{figure}[t]
  \centering
  \includegraphics[width=0.85\linewidth]{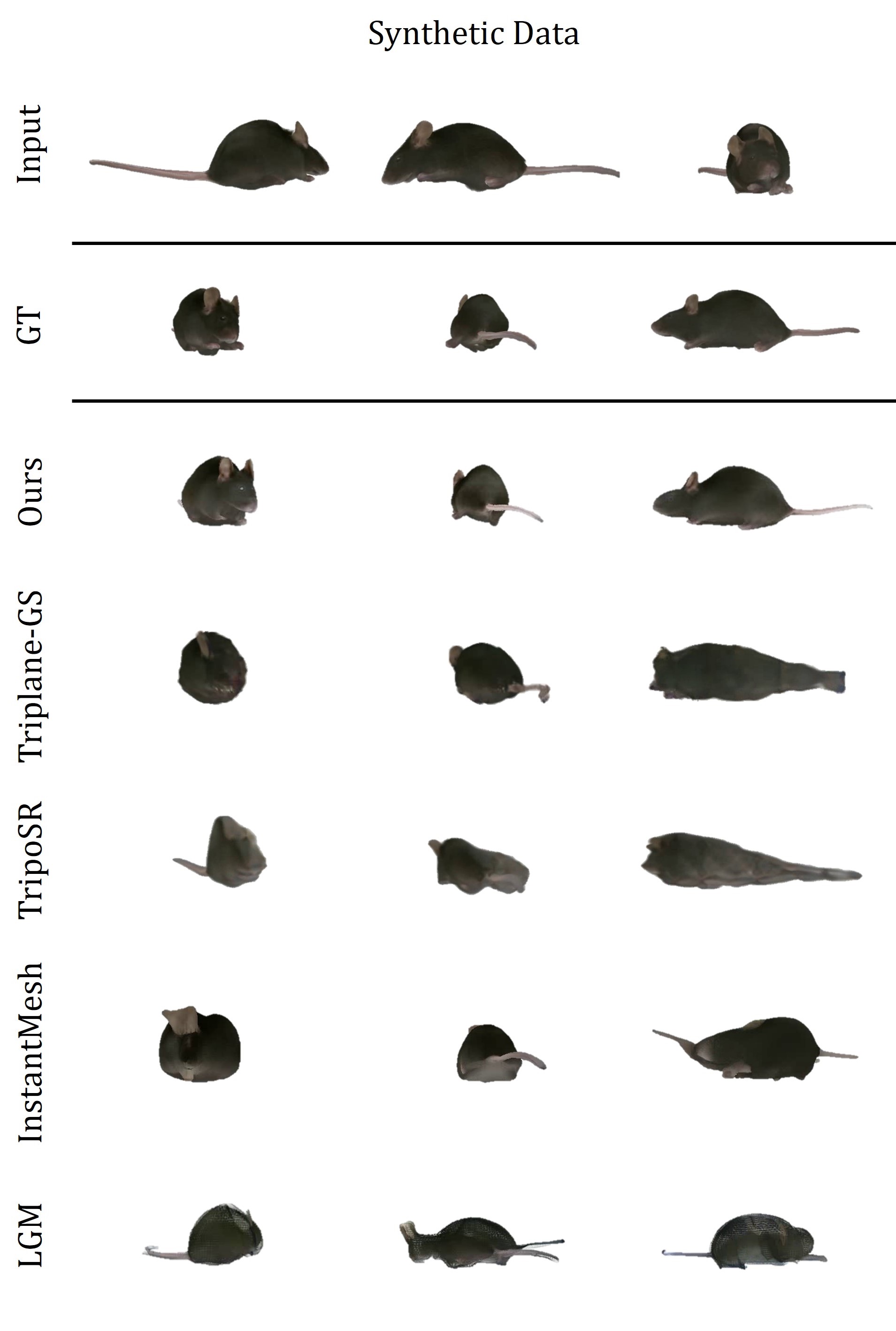}
  \caption{Qualitative comparison of novel view synthesis results on synthetic data. 
  MoReMouse produces coherent and anatomically plausible reconstructions, accurately capturing mouse posture and geometry, while baseline methods often exhibit structural distortions.}
  \label{fig:qual_results_synthetic}
\end{figure}

\begin{figure}[t]
  \centering
  \includegraphics[width=1.0\linewidth]{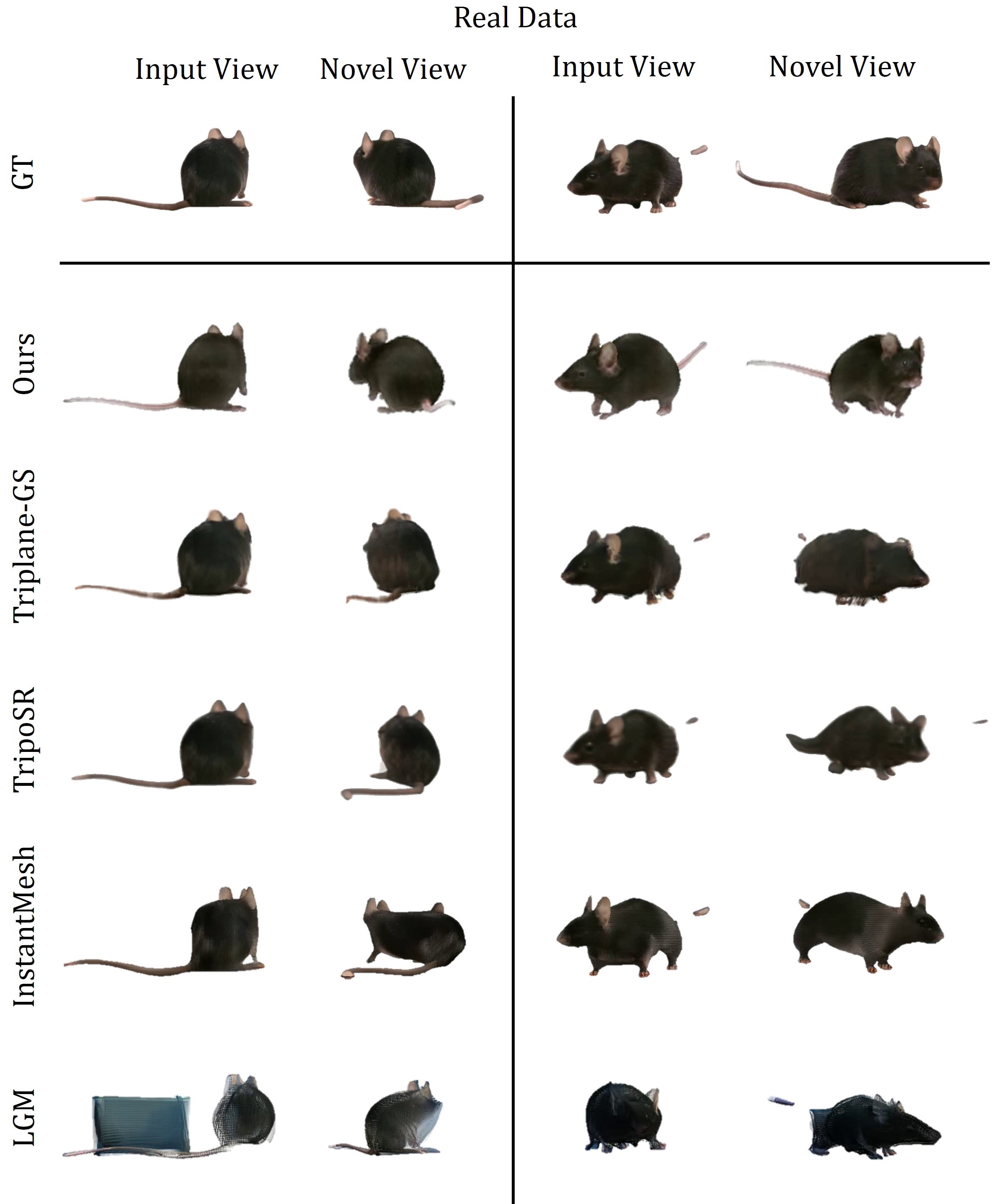}
  \caption{Qualitative comparison on real data. MoReMouse maintains consistent geometry and appearance, while baselines often show pose and shape artifacts. Occasional tail discontinuities arise from imperfect Track-Anything masks under reflections or occlusions; our model remains robust and recovers plausible tail shapes.}
  \label{fig:qual_results_real}
\end{figure}

\subsection{Experimental Results}

\paragraph{Quantitative Results.}
We report the evaluation results on the synthetic and real-world datasets in Table~\ref{tab:quant_combined}. For each metric, we highlight the top-3 performing methods using color shading, with darker tones indicating better performance.

Across all three metrics for novel view synthesis, MoReMouse consistently outperforms the baseline methods, demonstrating its ability to effectively infer plausible mouse geometry and pose from a single image.
We also observe that the quantitative metrics on the real-world dataset are generally lower than on the synthetic dataset, especially for PSNR. This is largely due to the camera-mouse configuration: in the real dataset, the cameras are fixed while the mouse moves, whereas MoReMouse and all baselines assume that the mouse is centered and aligned across input-output views. This mismatch introduces pixel-level alignment errors, reducing PSNR. However, the perceptual metrics SSIM and LPIPS remain high, indicating that MoReMouse still reconstructs plausible and correct mouse poses and shapes. This confirms that models trained on synthetic data can generalize to real-world mouse motion capture scenarios. Full tables of IoU metrics, expanded baseline comparisons, and runtime/memory usage are provided in the supplementary.


\paragraph{Qualitative Results.}
To visually compare MoReMouse with baseline methods, we select three representative frames from both the synthetic and real-world sequences. As shown in Fig.~\ref{fig:qual_results_synthetic} and Fig.~\ref{fig:qual_results_real}, MoReMouse generates mice with coherent appearance and physically plausible postures, accurately capturing tail and limb movements. These results remain stable even though each frame is inferred independently, showing smooth motion continuity. In contrast, baseline methods often exhibit structural distortions in novel views and inconsistent pose transitions across frames.

\subsection{Ablation Study}

\subsubsection{Comparison with Finetuned TripoSR}

\begin{figure}[t]
  \centering
  \includegraphics[width=1.0\linewidth]{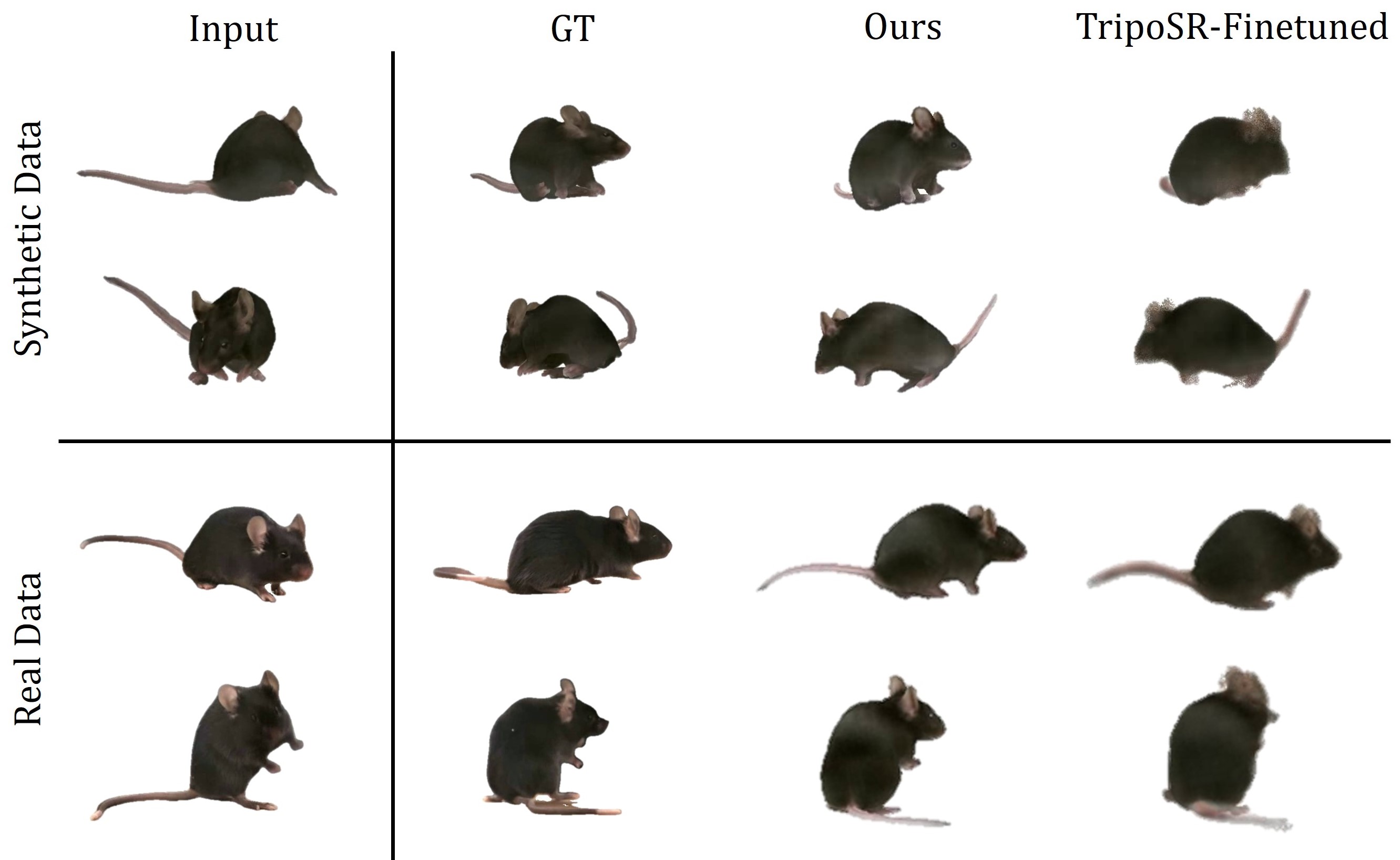}
  \caption{Qualitative comparison between our MoReMouse method and a finetuned TripoSR baseline. MoReMouse produces sharper and more anatomically coherent results. In contrast, TripoSR—based on volumetric NeRF rendering—tends to generate blurry boundaries and over-smoothed geometry, especially around fine structures like the ears.}
\label{fig:ablation_finetune}
\end{figure}

To further validate the benefits of our architectural design, we fine-tune TripoSR on our mouse dataset for 120 epochs with a learning rate of $1 \times 10^{-6}$. As shown in Table~\ref{tab:ours_vs_triposr-finetuned} and Fig.~\ref{fig:ablation_finetune}, the fine-tuned TripoSR shows noticeable improvement over its original version but still underperforms MoReMouse across most metrics. This indicates that, beyond data adaptation, our architectural choices—such as triplane encoding and geodesic embeddings—contribute significantly to reconstruction quality. We also fine-tune Triplane-GS and report its performance in the supplementary, which reveals reconstruction artifacts particularly around limbs and the tail.

\begin{table}[h]
\centering
\renewcommand{\arraystretch}{1.2}
\setlength{\tabcolsep}{4pt}
\begin{tabular}{llccc}
\toprule
\textbf{Dataset} & \textbf{Method} & \textbf{PSNR} $\uparrow$ & \textbf{SSIM} $\uparrow$ & \textbf{LPIPS} $\downarrow$ \\
\midrule
\multirow{2}{*}{\textit{synthetic}} 
& Ours & \cellcolor{cyan!15}22.027 & 0.9660 & \cellcolor{cyan!15}0.05279 \\
& TripoSR-Tuned & 21.996 & \cellcolor{cyan!15}0.9676 & 0.06333 \\
\midrule
\multirow{2}{*}{\textit{real}} 
& Ours & \cellcolor{cyan!15}18.422 & \cellcolor{cyan!15}0.9478 & \cellcolor{cyan!15}0.08674 \\
& TripoSR-Tuned & 18.162 & 0.9461 & 0.09354 \\
\bottomrule
\end{tabular}
\caption{Comparison between MoReMouse and TripoSR finetuned on mouse data.}
\label{tab:ours_vs_triposr-finetuned}
\end{table}

\subsubsection{Effect of Feature Embedding}

\begin{figure}[t]
  \centering
  \includegraphics[width=1.0\linewidth]{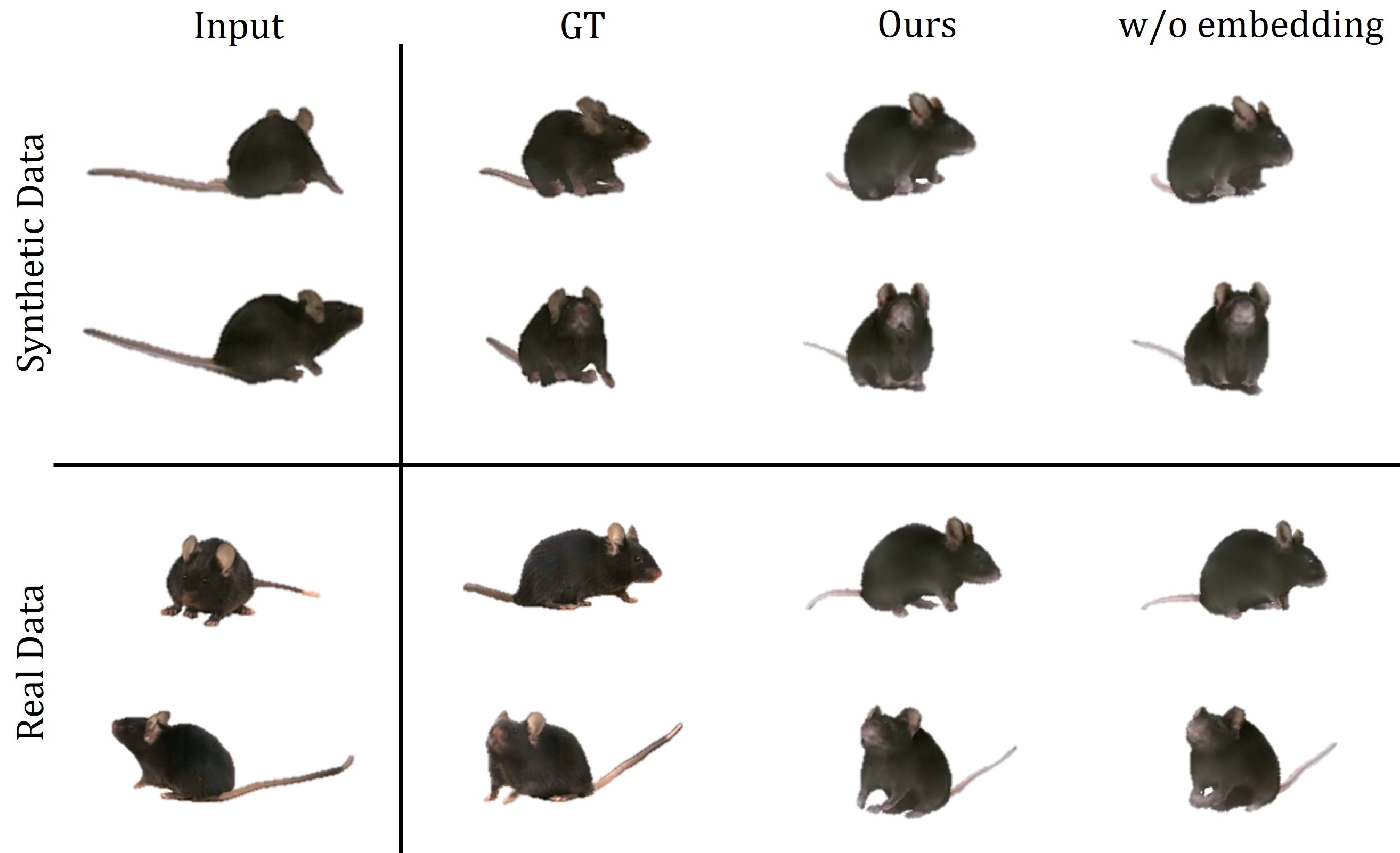}
  \caption{Qualitative comparison with and without geodesic feature embedding. The full model produces sharper and more consistent geometry, especially around limbs (rows 1, 3, 4) and snout texture (row 2). Embeddings help guide fine-grained surface details and improve anatomical accuracy.}
  \label{fig:ablation_embedding}
\end{figure}

We evaluate the effectiveness of the geodesic feature embedding module by training a variant of MoReMouse without this component. As shown in Table~\ref{tab:ablation_embedding} and visualized in Fig.~\ref{fig:ablation_embedding}, removing the embedding leads to noticeably degraded reconstruction quality, particularly around fine-scale structures.

Without embedding supervision, the model struggles to maintain surface consistency in anatomically delicate regions such as limbs and tails. As evident in the first, third, and fourth rows of Fig.~\ref{fig:ablation_embedding}, the full model produces sharper and more accurate reconstructions of the paws and extremities. In the second row, the embedded model also better preserves subtle texture details around the snout.These observations support our design choice to use surface-based semantic priors. The geodesic embedding encourages coherent geometry and finer detail recovery.

\begin{table}[h]
\centering
\renewcommand{\arraystretch}{1.2}
\setlength{\tabcolsep}{4pt}
\begin{tabular}{llccc}
\toprule
\textbf{Dataset} & \textbf{Method} & \textbf{PSNR} $\uparrow$ & \textbf{SSIM} $\uparrow$ & \textbf{LPIPS} $\downarrow$ \\
\midrule
\multirow{2}{*}{\textit{synthetic}} 
& full model & \cellcolor{cyan!15}22.027 & \cellcolor{cyan!15}0.9660 & \cellcolor{cyan!15}0.05279 \\
& w/o embedding & 21.805 & 0.9655 & 0.05467 \\
\midrule
\multirow{2}{*}{\textit{real}} 
& full model & \cellcolor{cyan!15}18.422 & \cellcolor{cyan!15}0.9478 & \cellcolor{cyan!15}0.08674 \\
& w/o embedding & 18.250 & 0.9469 & 0.08767 \\
\bottomrule
\end{tabular}
\caption{Ablation study on the effect of feature embedding. Removing embedding results in lower reconstruction quality, especially in perceptual similarity.}
\label{tab:ablation_embedding}
\end{table}






\section{Discussion}

\paragraph{Summary} In this paper, we present \textbf{MoReMouse}, the first monocular dense 3D reconstruction model tailored for laboratory mice. Unlike existing single-view 3D methods designed for general objects or humans, MoReMouse addresses the unique challenges posed by small animals, including non-rigid deformation, low surface texture, and limited training data. 
Our approach integrates transformer-based triplane representations, a Gaussian avatar modeling pipeline, and geodesic surface feature embeddings to achieve robust and high-fidelity reconstructions from a single image. We demonstrate strong performance across both synthetic and real datasets, outperforming state-of-the-art reconstruction baselines in terms of visual quality and perceptual consistency. Ablation studies further validate the architectural contributions of our model beyond dataset-specific priors. We hope that MoReMouse can serve as a foundation for future research in computational ethology and animal modeling, promoting more precise, efficient, and scalable behavioral analysis tools for biomedical research. 

\paragraph{Limitations.}
While MoReMouse offers a promising solution to monocular mouse reconstruction, several limitations remain that warrant future investigation. 
(1) \textit{Limited training diversity.} Our training data is derived from a single video sequence of a single mouse under consistent lighting and appearance. This limits the generalization of the model to other mice with varying fur color, size, or under different lighting conditions. 
(2) \textit{Pose tracking in global coordinates.} The current system assumes that the mouse is always centered in a canonical scene space. This design limits our ability to track full 3D poses in a real-world coordinate system. 

\section*{Ethical Statement} 
Please refer to supplementary file for Ethical Statements. 

\section*{Acknowledgements}
This work was supported by the National Natural Science Foundation of China (No. 62125107) and the Shuimu Tsinghua Scholar Program (2024SM324).

\bibliography{aaai2026}

@article{mathis2018deeplabcut,
  title={DeepLabCut: markerless pose estimation of user-defined body parts with deep learning},
  author={Mathis, Alexander and Mamidanna, Pranav and Cury, Kevin M and Abe, Taiga and Murthy, Venkatesh N and Mathis, Mackenzie Weygandt and Bethge, Matthias},
  journal={Nature neuroscience},
  volume={21},
  number={9},
  pages={1281--1289},
  year={2018},
  publisher={Nature Publishing Group US New York}
}

@article{pereira2022sleap,
  title={SLEAP: A deep learning system for multi-animal pose tracking},
  author={Pereira, Talmo D and Tabris, Nathaniel and Matsliah, Arie and Turner, David M and Li, Junyu and Ravindranath, Shruthi and Papadoyannis, Eleni S and Normand, Edna and Deutsch, David S and Wang, Z Yan and others},
  journal={Nature methods},
  volume={19},
  number={4},
  pages={486--495},
  year={2022},
  publisher={Nature Publishing Group US New York}
}

@article{tochilkin2024triposr,
  title={Triposr: Fast 3d object reconstruction from a single image},
  author={Tochilkin, Dmitry and Pankratz, David and Liu, Zexiang and Huang, Zixuan and Letts, Adam and Li, Yangguang and Liang, Ding and Laforte, Christian and Jampani, Varun and Cao, Yan-Pei},
  journal={arXiv preprint arXiv:2403.02151},
  year={2024}
}

@inproceedings{deitke2023objaverse,
  title={Objaverse: A universe of annotated 3d objects},
  author={Deitke, Matt and Schwenk, Dustin and Salvador, Jordi and Weihs, Luca and Michel, Oscar and VanderBilt, Eli and Schmidt, Ludwig and Ehsani, Kiana and Kembhavi, Aniruddha and Farhadi, Ali},
  booktitle={Proceedings of the IEEE/CVF conference on computer vision and pattern recognition},
  pages={13142--13153},
  year={2023}
}

@article{hong2023lrm,
  title={Lrm: Large reconstruction model for single image to 3d},
  author={Hong, Yicong and Zhang, Kai and Gu, Jiuxiang and Bi, Sai and Zhou, Yang and Liu, Difan and Liu, Feng and Sunkavalli, Kalyan and Bui, Trung and Tan, Hao},
  journal={arXiv preprint arXiv:2311.04400},
  year={2023}
}

@inproceedings{zou2024triplane,
  title={Triplane meets gaussian splatting: Fast and generalizable single-view 3d reconstruction with transformers},
  author={Zou, Zi-Xin and Yu, Zhipeng and Guo, Yuan-Chen and Li, Yangguang and Liang, Ding and Cao, Yan-Pei and Zhang, Song-Hai},
  booktitle={Proceedings of the IEEE/CVF conference on computer vision and pattern recognition},
  pages={10324--10335},
  year={2024}
}

@inproceedings{tang2024lgm,
  title={Lgm: Large multi-view gaussian model for high-resolution 3d content creation},
  author={Tang, Jiaxiang and Chen, Zhaoxi and Chen, Xiaokang and Wang, Tengfei and Zeng, Gang and Liu, Ziwei},
  booktitle={European Conference on Computer Vision},
  pages={1--18},
  year={2024},
  organization={Springer}
}

@article{xu2024instantmesh,
  title={Instantmesh: Efficient 3d mesh generation from a single image with sparse-view large reconstruction models},
  author={Xu, Jiale and Cheng, Weihao and Gao, Yiming and Wang, Xintao and Gao, Shenghua and Shan, Ying},
  journal={arXiv preprint arXiv:2404.07191},
  year={2024}
}

@article{mildenhall2021nerf,
  title={Nerf: Representing scenes as neural radiance fields for view synthesis},
  author={Mildenhall, Ben and Srinivasan, Pratul P and Tancik, Matthew and Barron, Jonathan T and Ramamoorthi, Ravi and Ng, Ren},
  journal={Communications of the ACM},
  volume={65},
  number={1},
  pages={99--106},
  year={2021},
  publisher={ACM New York, NY, USA}
}

@article{shen2021deep,
  title={Deep marching tetrahedra: a hybrid representation for high-resolution 3d shape synthesis},
  author={Shen, Tianchang and Gao, Jun and Yin, Kangxue and Liu, Ming-Yu and Fidler, Sanja},
  journal={Advances in Neural Information Processing Systems},
  volume={34},
  pages={6087--6101},
  year={2021}
}

@article{an2023three,
  title={Three-dimensional surface motion capture of multiple freely moving pigs using MAMMAL},
  author={An, Liang and Ren, Jilong and Yu, Tao and Hai, Tang and Jia, Yichang and Liu, Yebin},
  journal={Nature Communications},
  volume={14},
  number={1},
  pages={7727},
  year={2023},
  publisher={Nature Publishing Group UK London}
}

@inproceedings{li2024animatable,
  title={Animatable gaussians: Learning pose-dependent gaussian maps for high-fidelity human avatar modeling},
  author={Li, Zhe and Zheng, Zerong and Wang, Lizhen and Liu, Yebin},
  booktitle={Proceedings of the IEEE/CVF conference on computer vision and pattern recognition},
  pages={19711--19722},
  year={2024}
}

@article{bohnslav2023armo,
  title={ArMo: An Articulated Mesh Approach for Mouse 3D Reconstruction},
  author={Bohnslav, James P and Osman, Mohammed Abdal Monium and Jaggi, Akshay and Soares, Sofia and Weinreb, Caleb and Datta, Sandeep Robert and Harvey, Christopher D},
  journal={bioRxiv},
  pages={2023--02},
  year={2023},
  publisher={Cold Spring Harbor Laboratory}
}

@article{dunn2021geometric,
  title={Geometric deep learning enables 3D kinematic profiling across species and environments},
  author={Dunn, Timothy W and Marshall, Jesse D and Severson, Kyle S and Aldarondo, Diego E and Hildebrand, David GC and Chettih, Selmaan N and Wang, William L and Gellis, Amanda J and Carlson, David E and Aronov, Dmitriy and others},
  journal={Nature methods},
  volume={18},
  number={5},
  pages={564--573},
  year={2021},
  publisher={Nature Publishing Group US New York}
}

@article{35_nath2019using,
  title={Using DeepLabCut for 3D markerless pose estimation across species and behaviors},
  author={Nath, Tanmay and Mathis, Alexander and Chen, An Chi and Patel, Amir and Bethge, Matthias and Mathis, Mackenzie Weygandt},
  journal={Nature Protocols},
  volume={14},
  number={7},
  pages={2152--2176},
  year={2019},
  publisher={Nature Publishing Group UK London}
}

@article{36_karashchuk2021anipose,
  title={Anipose: a toolkit for robust markerless 3D pose estimation},
  author={Karashchuk, Pierre and Rupp, Katie L and Dickinson, Evyn S and Walling-Bell, Sarah and Sanders, Elischa and Azim, Eiman and Brunton, Bingni W and Tuthill, John C},
  journal={Cell Reports},
  volume={36},
  number={13},
  year={2021},
  publisher={Elsevier}
}

@article{klibaite2025mapping,
  title={Mapping the landscape of social behavior},
  author={Klibaite, Ugne and Li, Tianqing and Aldarondo, Diego and Akoad, Jumana F and {\"O}lveczky, Bence P and Dunn, Timothy W},
  journal={Cell},
  volume={188},
  number={8},
  pages={2249--2266},
  year={2025},
  publisher={Elsevier}
}

@article{han2024multi,
  title={Multi-animal 3D social pose estimation, identification and behaviour embedding with a few-shot learning framework},
  author={Han, Yaning and Chen, Ke and Wang, Yunke and Liu, Wenhao and Wang, Zhouwei and Wang, Xiaojing and Han, Chuanliang and Liao, Jiahui and Huang, Kang and Cai, Shengyuan and others},
  journal={Nature Machine Intelligence},
  pages={1--14},
  year={2024},
  publisher={Nature Publishing Group UK London}
}

@article{49_bolanos2021three,
  title={A three-dimensional virtual mouse generates synthetic training data for behavioral analysis},
  author={Bola{\~n}os, Luis A and Xiao, Dongsheng and Ford, Nancy L and LeDue, Jeff M and Gupta, Pankaj K and Doebeli, Carlos and Hu, Hao and Rhodin, Helge and Murphy, Timothy H},
  journal={Nature Methods},
  volume={18},
  number={4},
  pages={378--381},
  year={2021},
  publisher={Nature Publishing Group US New York}
}

@inproceedings{chan2022efficient,
	title        = {Efficient geometry-aware 3D generative adversarial networks},
	author       = {Chan, Eric R and Lin, Connor Z and Chan, Matthew A and Nagano, Koki and Pan, Boxiao and De Mello, Shalini and Gallo, Orazio and Guibas, Leonidas J and Tremblay, Jonathan and Khamis, Sameh and others},
	year         = 2022,
	booktitle    = {Proceedings of the IEEE/CVF Conference on Computer Vision and Pattern Recognition},
	pages        = {16123--16133}
}

@inproceedings{radford2021learning,
	title        = {Learning transferable visual models from natural language supervision},
	author       = {Radford, Alec and Kim, Jong Wook and Hallacy, Chris and Ramesh, Aditya and Goh, Gabriel and Agarwal, Sandhini and Sastry, Girish and Askell, Amanda and Mishkin, Pamela and Clark, Jack and others},
	year         = 2021,
	booktitle    = {International conference on machine learning},
	pages        = {8748--8763},
	organization = {PMLR}
}

@inproceedings{wu2023omniobject3d,
  title={Omniobject3d: Large-vocabulary 3d object dataset for realistic perception, reconstruction and generation},
  author={Wu, Tong and Zhang, Jiarui and Fu, Xiao and Wang, Yuxin and Ren, Jiawei and Pan, Liang and Wu, Wayne and Yang, Lei and Wang, Jiaqi and Qian, Chen and others},
  booktitle={Proceedings of the IEEE/CVF Conference on Computer Vision and Pattern Recognition},
  pages={803--814},
  year={2023}
}

@article{kerbl2023gaussian,
  title={3d gaussian splatting for real-time radiance field rendering},
  author={Kerbl, Bernhard and Kopanas, Georgios and Leimk{\"u}hler, Thomas and Drettakis, George},
  journal={TOG},
  volume={42},
  number={4},
  pages={1--14},
  year={2023},
  publisher={ACM New York, NY, USA}
}

@article{Li2023Instant3DFT,
  title={Instant3D: Fast Text-to-3D with Sparse-View Generation and Large Reconstruction Model},
  author={Jiahao Li and Hao Tan and Kai Zhang and Zexiang Xu and Fujun Luan and Yinghao Xu and Yicong Hong and Kalyan Sunkavalli and Greg Shakhnarovich and Sai Bi},
  journal={ArXiv},
  year={2023},
  volume={abs/2311.06214},
  url={https://api.semanticscholar.org/CorpusID:265128529}
}

@article{gslrm2024,
    author={Zhang, Kai and Bi, Sai and Tan, Hao and Xiangli, Yuanbo and Zhao, Nanxuan 
      and Sunkavalli, Kalyan and Xu, Zexiang},
    title     = {GS-LRM: Large Reconstruction Model for 3D Gaussian Splatting},
    journal   = {arXiv},
    year      = {2024},
}

@article{zhang2024clay,
  title={CLAY: A Controllable Large-scale Generative Model for Creating High-quality 3D Assets},
  author={Zhang, Longwen and Wang, Ziyu and Zhang, Qixuan and Qiu, Qiwei and Pang, Anqi and Jiang, Haoran and Yang, Wei and Xu, Lan and Yu, Jingyi},
  journal={ACM Transactions on Graphics (TOG)},
  volume={43},
  number={4},
  pages={1--20},
  year={2024},
  publisher={ACM New York, NY, USA}
}

@article{sun2024dreamcraft3d++,
  title={DreamCraft3D++: Efficient Hierarchical 3D Generation with Multi-Plane Reconstruction Model},
  author={Sun, Jingxiang and Peng, Cheng and Shao, Ruizhi and Guo, Yuan-Chen and Zhao, Xiaochen and Li, Yangguang and Cao, Yanpei and Zhang, Bo and Liu, Yebin},
  journal={arXiv preprint arXiv:2410.12928},
  year={2024}
}

@article{aldarondo2024virtual,
  title={A virtual rodent predicts the structure of neural activity across behaviours},
  author={Aldarondo, Diego and Merel, Josh and Marshall, Jesse D and Hasenclever, Leonard and Klibaite, Ugne and Gellis, Amanda and Tassa, Yuval and Wayne, Greg and Botvinick, Matthew and {\"O}lveczky, Bence P},
  journal={Nature},
  volume={632},
  number={8025},
  pages={594--602},
  year={2024},
  publisher={Nature Publishing Group UK London}
}

@article{monsees2022estimation,
  title={Estimation of skeletal kinematics in freely moving rodents},
  author={Monsees, Arne and Voit, Kay-Michael and Wallace, Damian J and Sawinski, Juergen and Charyasz, Edyta and Scheffler, Klaus and Macke, Jakob H and Kerr, Jason ND},
  journal={Nature Methods},
  volume={19},
  number={11},
  pages={1500--1509},
  year={2022},
  publisher={Nature Publishing Group US New York}
}

@article{marshall2021pair,
  title={The pair-r24m dataset for multi-animal 3d pose estimation},
  author={Marshall, Jesse D and Klibaite, Ugne and Gellis, Amanda and Aldarondo, Diego E and {\"O}lveczky, Bence P and Dunn, Timothy W},
  journal={bioRxiv},
  pages={2021--11},
  year={2021},
  publisher={Cold Spring Harbor Laboratory}
}

@inproceedings{63_zuffi20173d,
  title={3D menagerie: Modeling the 3D shape and pose of animals},
  author={Zuffi, Silvia and Kanazawa, Angjoo and Jacobs, David W and Black, Michael J},
  booktitle={Proceedings of the IEEE Conference on Computer Vision and Pattern Recognition},
  pages={6365--6373},
  year={2017}
}

@inproceedings{xu2023animal3d,
  title={Animal3d: A comprehensive dataset of 3d animal pose and shape},
  author={Xu, Jiacong and Zhang, Yi and Peng, Jiawei and Ma, Wufei and Jesslen, Artur and Ji, Pengliang and Hu, Qixin and Zhang, Jiehua and Liu, Qihao and Wang, Jiahao and others},
  booktitle={Proceedings of the IEEE/CVF International Conference on Computer Vision},
  pages={9099--9109},
  year={2023}
}

@inproceedings{zuffi2019three,
  title={Three-D Safari: Learning to Estimate Zebra Pose, Shape, and Texture from Images" In the Wild"},
  author={Zuffi, Silvia and Kanazawa, Angjoo and Berger-Wolf, Tanya and Black, Michael J},
  booktitle={Proceedings of the IEEE/CVF International Conference on Computer Vision},
  pages={5359--5368},
  year={2019}
}

@inproceedings{lyu2025animer,
  title={AniMer: Animal Pose and Shape Estimation Using Family Aware Transformer},
  author={Lyu, Jin and Zhu, Tianyi and Gu, Yi and Lin, Li and Cheng, Pujin and Liu, Yebin and Tang, Xiaoying and An, Liang},
  booktitle={Proceedings of the Computer Vision and Pattern Recognition Conference},
  pages={17486--17496},
  year={2025}
}

@inproceedings{biggs2020wldo,
  title={Who left the dogs out? 3d animal reconstruction with expectation maximization in the loop},
  author={Biggs, Benjamin and Boyne, Oliver and Charles, James and Fitzgibbon, Andrew and Cipolla, Roberto},
  booktitle={European Conference on Computer Vision},
  pages={195--211},
  year={2020},
  organization={Springer}
}

@inproceedings{rueegg2022barc,
  title={Barc: Learning to regress 3d dog shape from images by exploiting breed information},
  author={Rueegg, Nadine and Zuffi, Silvia and Schindler, Konrad and Black, Michael J},
  booktitle={Proceedings of the IEEE/CVF Conference on Computer Vision and Pattern Recognition},
  pages={3876--3884},
  year={2022}
}

@inproceedings{Sabathier2024AnimalAR,
  title={Animal avatars: Reconstructing animatable 3D animals from casual videos},
  author={Sabathier, Remy and Mitra, Niloy J and Novotny, David},
  booktitle={European Conference on Computer Vision},
  pages={270--287},
  year={2024},
  organization={Springer}
}

@inproceedings{zuffi2024awol,
  title={AWOL: Analysis WithOut Synthesis Using Language},
  author={Zuffi, Silvia and Black, Michael J},
  booktitle={European Conference on Computer Vision},
  pages={1--19},
  year={2024},
  organization={Springer}
}

@inproceedings{yang2022banmo,
  title={Banmo: Building animatable 3d neural models from many casual videos},
  author={Yang, Gengshan and Vo, Minh and Neverova, Natalia and Ramanan, Deva and Vedaldi, Andrea and Joo, Hanbyul},
  booktitle={Proceedings of the IEEE/CVF Conference on Computer Vision and Pattern Recognition},
  pages={2863--2873},
  year={2022}
}

@inproceedings{lei2024gart,
  title={Gart: Gaussian articulated template models},
  author={Lei, Jiahui and Wang, Yufu and Pavlakos, Georgios and Liu, Lingjie and Daniilidis, Kostas},
  booktitle={Proceedings of the IEEE/CVF conference on computer vision and pattern recognition},
  pages={19876--19887},
  year={2024}
}

@inproceedings{wang2023styleavatar,
  title={Styleavatar: Real-time photo-realistic portrait avatar from a single video},
  author={Wang, Lizhen and Zhao, Xiaochen and Sun, Jingxiang and Zhang, Yuxiang and Zhang, Hongwen and Yu, Tao and Liu, Yebin},
  booktitle={ACM SIGGRAPH 2023 Conference Proceedings},
  pages={1--10},
  year={2023}
}

@inproceedings{wu2023magicpony,
  title={Magicpony: Learning articulated 3d animals in the wild},
  author={Wu, Shangzhe and Li, Ruining and Jakab, Tomas and Rupprecht, Christian and Vedaldi, Andrea},
  booktitle={Proceedings of the IEEE/CVF Conference on Computer Vision and Pattern Recognition},
  pages={8792--8802},
  year={2023}
}

@inproceedings{li2024learning,
  title={Learning the 3d fauna of the web},
  author={Li, Zizhang and Litvak, Dor and Li, Ruining and Zhang, Yunzhi and Jakab, Tomas and Rupprecht, Christian and Wu, Shangzhe and Vedaldi, Andrea and Wu, Jiajun},
  booktitle={Proceedings of the IEEE/CVF Conference on Computer Vision and Pattern Recognition},
  pages={9752--9762},
  year={2024}
}

@inproceedings{shamai2017geodesic,
  title={Geodesic distance descriptors},
  author={Shamai, Gil and Kimmel, Ron},
  booktitle={Proceedings of the IEEE Conference on Computer Vision and Pattern Recognition},
  pages={6410--6418},
  year={2017}
}

@inproceedings{xiang2025structured,
  title={Structured 3d latents for scalable and versatile 3d generation},
  author={Xiang, Jianfeng and Lv, Zelong and Xu, Sicheng and Deng, Yu and Wang, Ruicheng and Zhang, Bowen and Chen, Dong and Tong, Xin and Yang, Jiaolong},
  booktitle={Proceedings of the Computer Vision and Pattern Recognition Conference},
  pages={21469--21480},
  year={2025}
}

@article{hu2025simulating,
  title={Simulating the real world: A unified survey of multimodal generative models},
  author={Hu, Yuqi and Wang, Longguang and Liu, Xian and Chen, Ling-Hao and Guo, Yuwei and Shi, Yukai and Liu, Ce and Rao, Anyi and Wang, Zeyu and Xiong, Hui},
  journal={arXiv preprint arXiv:2503.04641},
  year={2025}
}

@article{zhang20233dshape2vecset,
  title={3dshape2vecset: A 3d shape representation for neural fields and generative diffusion models},
  author={Zhang, Biao and Tang, Jiapeng and Niessner, Matthias and Wonka, Peter},
  journal={ACM Transactions On Graphics (TOG)},
  volume={42},
  number={4},
  pages={1--16},
  year={2023},
  publisher={ACM New York, NY, USA}
}

@inproceedings{li2024craftsman3d,
  title={CraftsMan3D: High-fidelity Mesh Generation with 3D Native Diffusion and Interactive Geometry Refiner},
  author={Li, Weiyu and Liu, Jiarui and Yan, Hongyu and Chen, Rui and Liang, Yixun and Chen, Xuelin and Tan, Ping and Long, Xiaoxiao},
  booktitle={Proceedings of the Computer Vision and Pattern Recognition Conference},
  pages={5307--5317},
  year={2025}
}

@article{gupta20233dgen,
  title={3dgen: Triplane latent diffusion for textured mesh generation},
  author={Gupta, Anchit and Xiong, Wenhan and Nie, Yixin and Jones, Ian and O{\u{g}}uz, Barlas},
  journal={arXiv preprint arXiv:2303.05371},
  year={2023}
}

@article{xiong2024octfusion,
  title={Octfusion: Octree-based diffusion models for 3d shape generation},
  author={Xiong, Bojun and Wei, Si-Tong and Zheng, Xin-Yang and Cao, Yan-Pei and Lian, Zhouhui and Wang, Peng-Shuai},
  journal={arXiv preprint arXiv:2408.14732},
  year={2024}
}

@misc{yang2023trackanythingsegmentmeets,
      title={Track Anything: Segment Anything Meets Videos}, 
      author={Jinyu Yang and Mingqi Gao and Zhe Li and Shang Gao and Fangjing Wang and Feng Zheng},
      year={2023},
      eprint={2304.11968},
      archivePrefix={arXiv},
      primaryClass={cs.CV},
      url={https://arxiv.org/abs/2304.11968}, 
}

@article{maaten2008visualizing,
  title={Visualizing data using t-SNE},
  author={Maaten, Laurens van der and Hinton, Geoffrey},
  journal={Journal of machine learning research},
  volume={9},
  number={Nov},
  pages={2579--2605},
  year={2008}
}

\clearpage
\appendix
\setcounter{secnumdepth}{2}
\setcounter{figure}{0}
\setcounter{table}{0}
\renewcommand{\thesection}{A\arabic{section}}
\renewcommand{\thetable}{A\arabic{table}}
\renewcommand{\thefigure}{A\arabic{figure}}
\section*{Appendix}

\begin{figure*}[h]
    \centering
    \includegraphics[width=\linewidth]{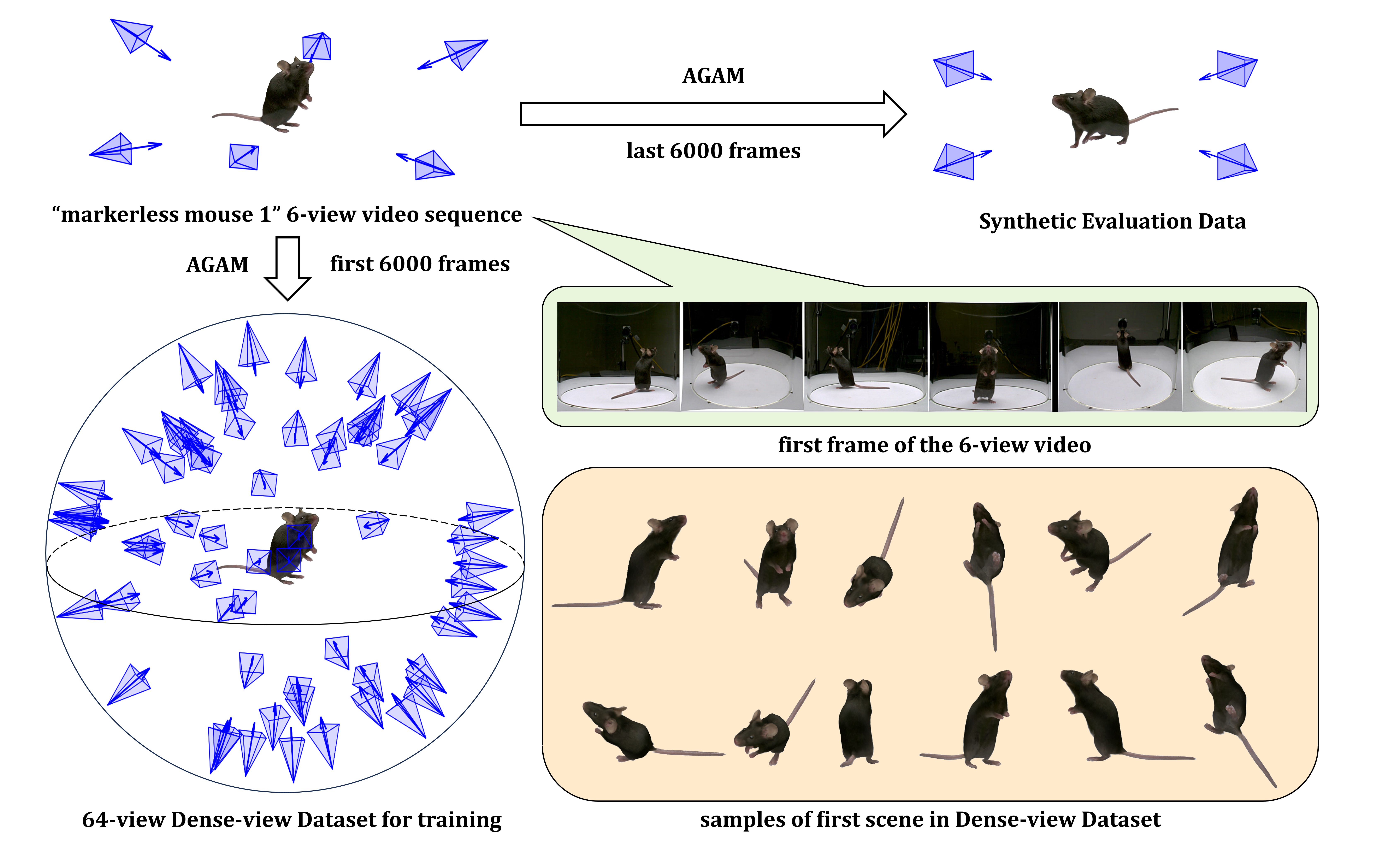}
    \caption{Pipeline for constructing the \textit{Dense-view Dataset} and synthetic evaluation data. 
    The first 6{,}000 frames of the ``markerless\_mouse\_1'' 6-view sequence are used to build the dense-view dataset for training, while the last 6{,}000 frames are reserved for generating the synthetic evaluation set. 
    For each training frame, AGAM is rendered from 64 randomly sampled viewpoints on a sphere of radius $2.22$, 
    resulting in $12{,}000$ multi-view scenes.}
    \label{fig:construction_dataset}
\end{figure*}

\section{Ethical Statement} 
All procedures involving mice in this study were non-invasive and observational, involving only video recording of natural behaviors without any physical intervention or stress to the animals. The experiments were conducted in accordance with internationally recognized guidelines for laboratory animal care. Since the study involved only observational methods, ethical approval was not required by the relevant institutional animal care and local ethics authority. The mice were housed under standard conditions in compliance with local animal welfare regulations, and all efforts were made to minimize disturbance during filming. The study adhered to the 3Rs principle (Replacement, Reduction, Refinement) in animal research.


\section{Dataset Construction}

\subsection{Dense-view Dataset Generation}
As illustrated in Fig.~\ref{fig:construction_dataset}, we construct our \textit{Dense-view Dataset} using the first 6{,}000 frames of the ``markerless\_mouse\_1'' video sequence. For each frame, we sample two independent sets of $64$ viewpoints, producing $12{,}000$ multi-view scenes in total, each containing $64$ images.

For viewpoint sampling, we first align the local coordinate origin of the generated Animatable Gaussian Avatar of Mouse (AGAM)---given by the origin of its internally driven mesh model---to the world coordinate origin. The Gaussian mouse model is uniformly scaled 
down by a factor of $180$ so that the mouse fits inside a unit sphere. We then randomly sample camera centers on a sphere of radius $2.22$, allowing for both top-down and bottom-up observations, so our dataset can accommodate to scenarios where mice are recorded from above or through a transparent floor. 
The rendered images have a resolution of $800\times800$ pixels, and the FoV is fixed at $29.86^{\circ}$.

\subsection{Synthetic Evaluation Dataset Generation}
We use the last 6{,}000 frames from the ``markerless\_mouse\_1'' sequence to construct a \textit{Synthetic Evaluation Dataset} consisting of four orthogonal viewpoints (front, back, left, and right). 
Following a similar placement strategy as in the \textit{Dense-view Dataset}, we align the local coordinate origin of the Gaussian mouse model with the world coordinate origin and uniformly scale it down by a factor of $180$ so that it fits within a unit sphere. 
The four cameras are positioned on a sphere of radius $2.22$ with optical axes pointing toward the scene origin, evenly spaced in azimuth by $90^\circ$. 
The rendered images use a fixed FoV of $29.86^{\circ}$ and have a resolution of $800\times800$. 

\begin{figure}[!h]
    \centering
    \includegraphics[width=0.9\linewidth]{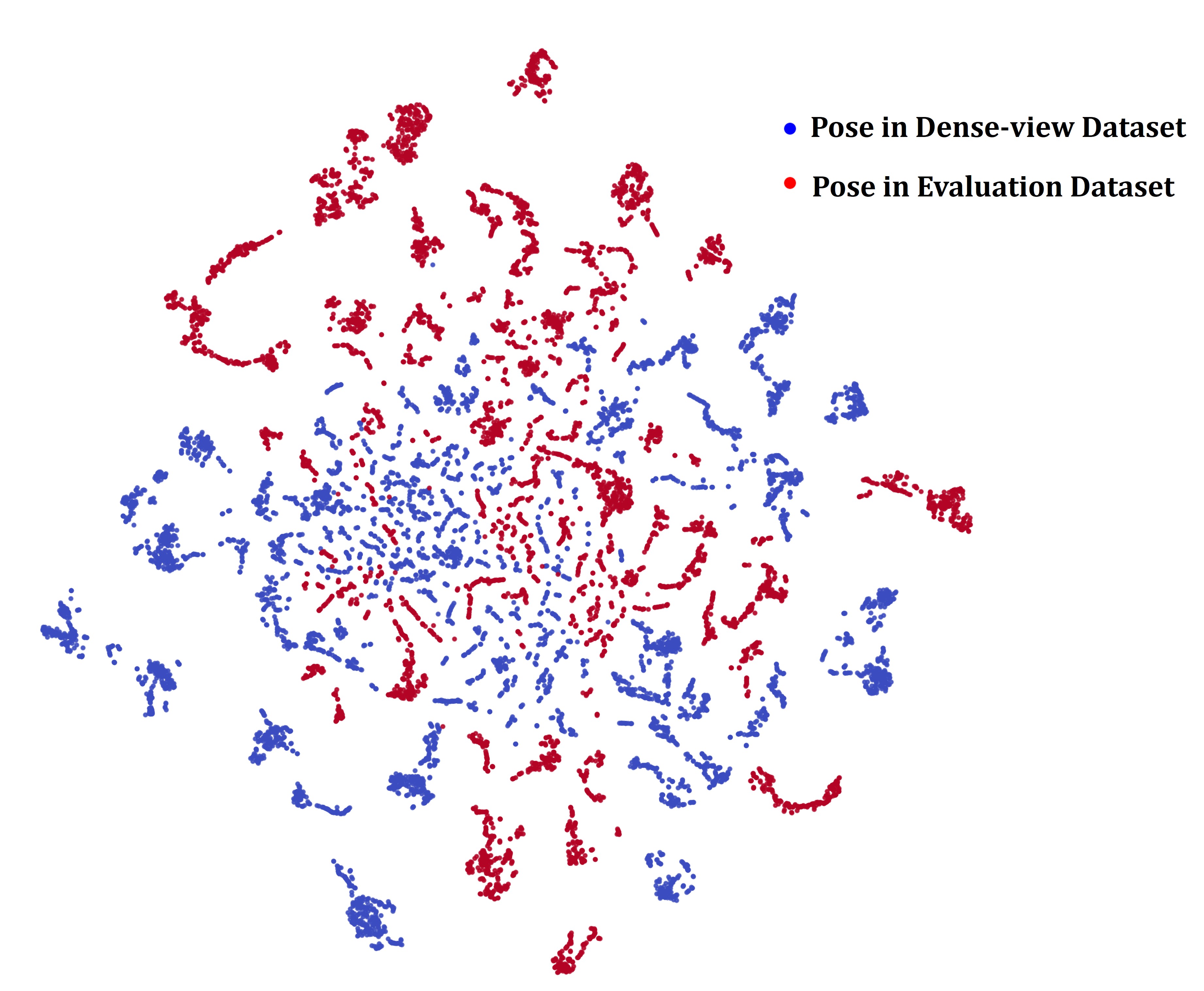}
    \caption{\textbf{t-SNE visualization of mouse poses.} The visualization shows that the training and testing poses are separated in the embedding space. Blue points are poses from Dense-view training dataset, red points are poses from the evaluation set.}
    \label{fig:tsne_pose_distribution}
\end{figure}

To verify that the poses in the evaluation set differ from those in the training set, we visualize the pose distributions of the \textit{Dense-view Dataset} and \textit{Synthetic Evaluation Dataset} using t-SNE~\cite{maaten2008visualizing}, as shown in Fig.~\ref{fig:tsne_pose_distribution}. Note that the evaluation set contains a large number of unseen poses, ensuring that the reported performance reflects the model’s ability to generalize to novel poses.

\subsection{Real Captured Data}
\begin{figure}[!h]
    \centering
    \includegraphics[width=\linewidth]{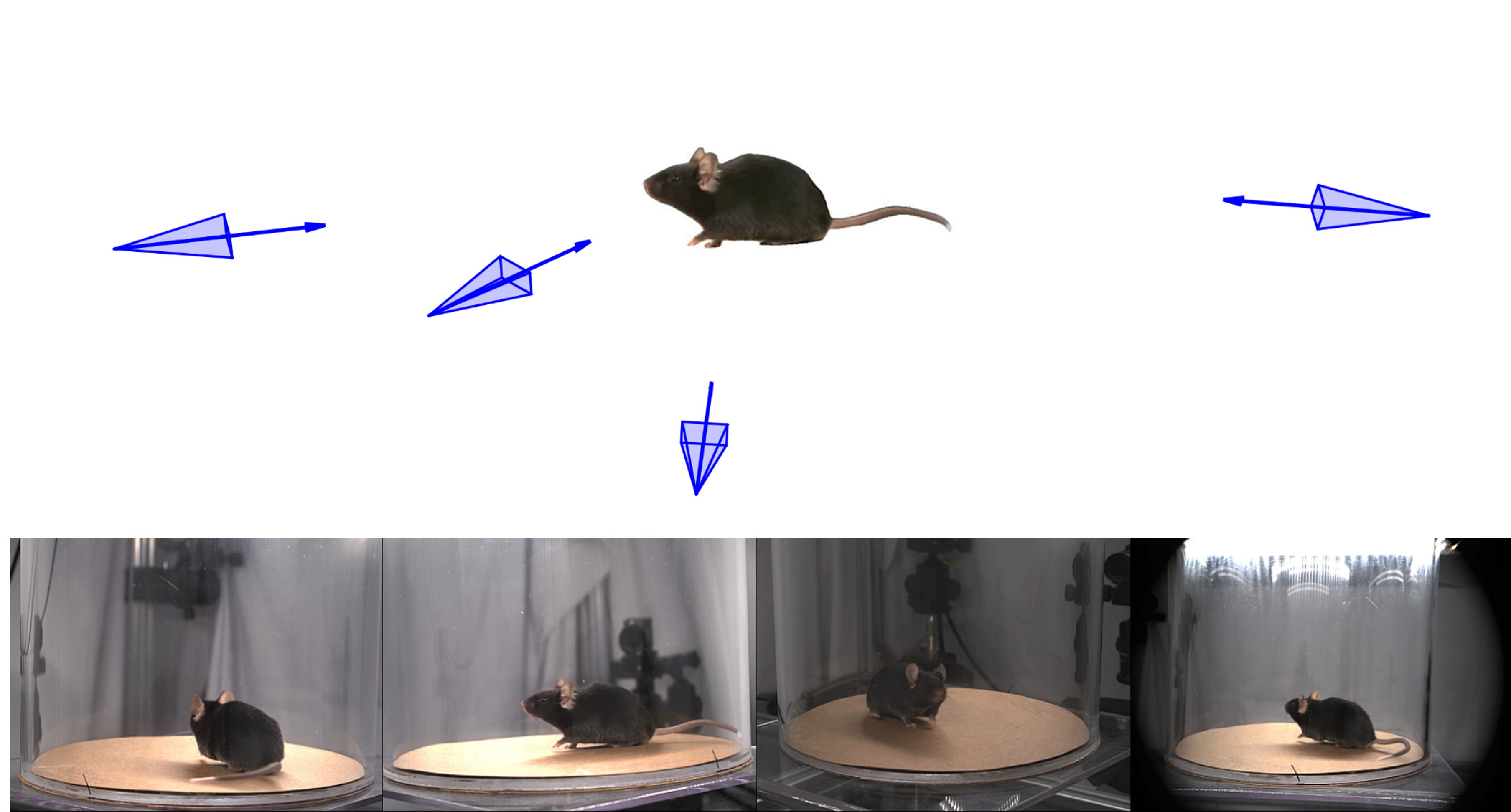}
    \caption{\textbf{Real-world dataset acquisition setup.} 
    Four PointGrey industrial cameras are arranged to capture a freely moving C57BL/6 mice. 
    This dataset enables the evaluation of cross-domain generalization from synthetic to real-world conditions.}
    \label{fig:construction_real_dataset}
\end{figure}

As shown in Fig.~\ref{fig:construction_real_dataset}, we collect a real-world multi-view dataset using PointGrey industrial cameras. 
The capture setup consists of four synchronized cameras positioned around a freely moving C57BL/6 mouse. The mouse is placed inside a transparent cylinder with a diameter of 20cm and a height of 20cm.   
Each camera records at a resolution of $2048\times1500$ pixels, with a horizontal field-of-view (FoV) of $22.34^{\circ}$ and a vertical FoV of $16.44^{\circ}$. 
This real-world dataset serves to evaluate the generalization of our model from synthetic training data to realistic environments. Before feeding the images to the network, all mice are cropped out with background removed. 

\begin{figure*}[h]
    \centering
    \includegraphics[width=\linewidth]{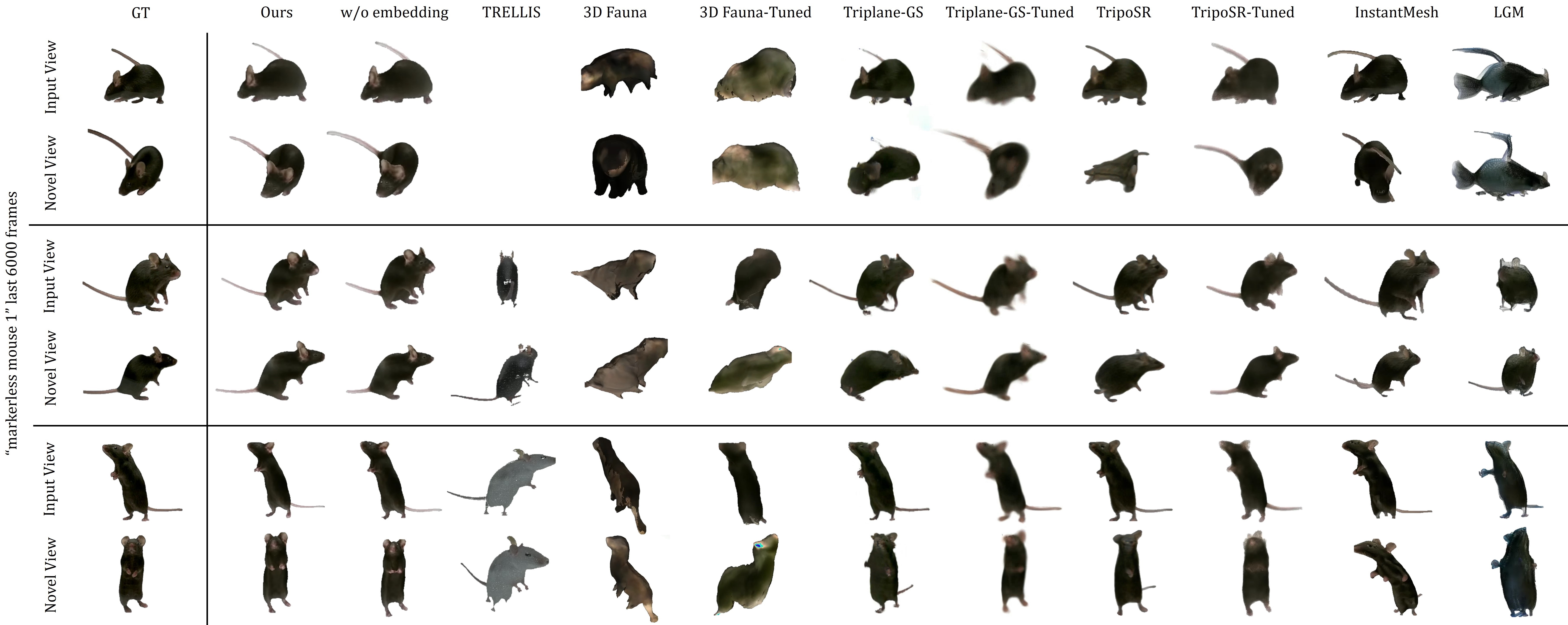}
    \caption{\textbf{Qualitative comparison on the original 6-view ``markerless\_mouse\_1'' dataset.}  
    Our method produces more geometrically accurate and visually consistent reconstructions compared to other baselines. 
    We also observe that \textbf{TRELLIS} occasionally outputs nearly blank images, indicating instability when handling mouse poses.}
    \label{fig:large_compare_ori}
\end{figure*}

\begin{figure*}[h]
    \centering
    \includegraphics[width=\linewidth]{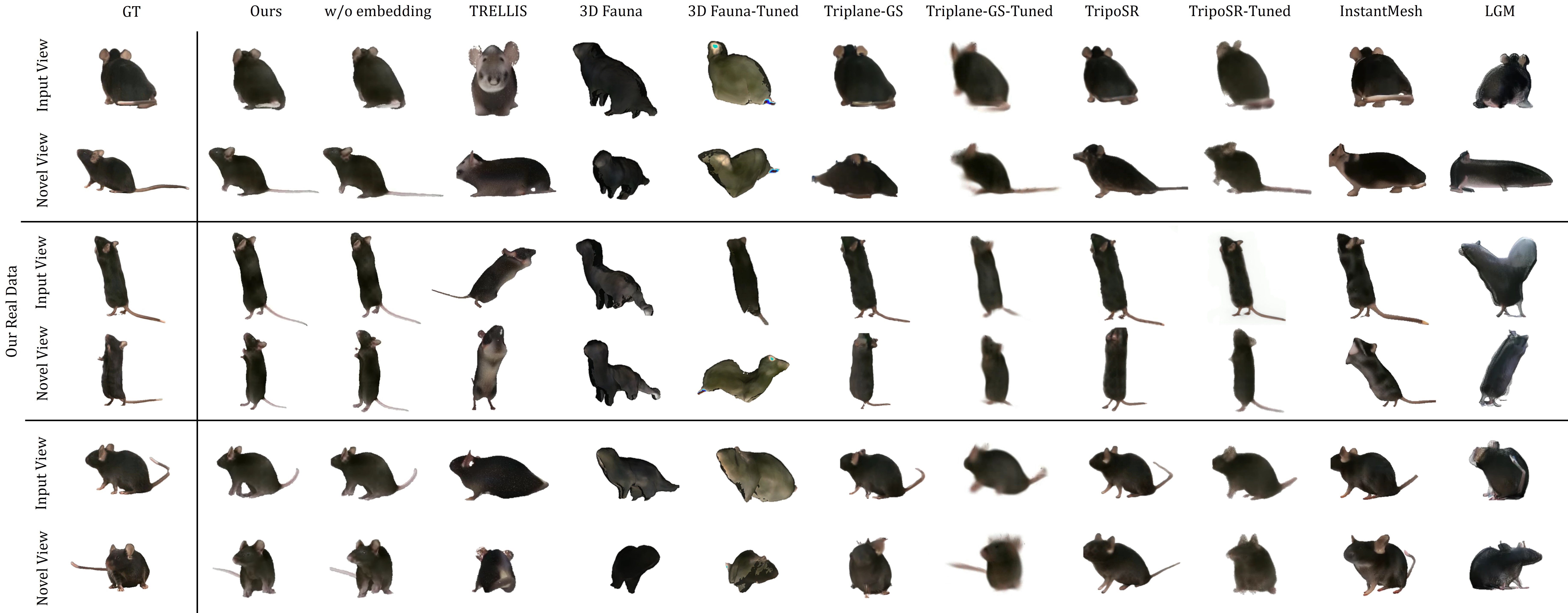}
    \caption{\textbf{Qualitative comparison on the real-world multi-view dataset.}  
    Our method reconstructs anatomically coherent geometry and consistent appearance across views, 
    whereas competing methods often produce structural distortions or inconsistent textures.}
    \label{fig:large_compare_real}
\end{figure*}

\section{More Experimental Results}

In this section, we present additional experimental results beyond those reported in the main paper. 
In addition to the four baseline methods compared in the main text  
(\textbf{TripoSR}~\cite{tochilkin2024triposr},  
\textbf{Triplane-GS}~\cite{zou2024triplane},  
\textbf{LGM}~\cite{tang2024lgm},  
and \textbf{InstantMesh}~\cite{xu2024instantmesh}),  
we further compare our method against two recent reconstruction approaches:  
\begin{itemize}
    \item \textbf{3D Fauna}~\cite{li2024learning}: A template-free method for learning 3D reconstruction of animals from online monocular images.  
    \item \textbf{TRELLIS}~\cite{xiang2025structured}: A structured generative model designed to capture fine-grained articulations through a hierarchical representation.  
\end{itemize}

We also report a surface geometry metric based on the IoU between the rendered reconstruction and multi-view visual-hull masks, offering a direct measure of geometric alignment.

To further examine the behavior of Triplane-GS on our task, we fine-tune it on our mouse dataset for 200 epochs using a learning rate of $1\times10^{-6}$. However, the resulting Gaussian point clouds remain blurry and lack sharp geometric boundaries, making it difficult to capture fine structures such as limbs and tails.

We additionally adapt 3D-Fauna’s pan-category deformable model to our 6-view mouse video dataset via fine-tuning. During training, each iteration samples data with a probability of $0.6$ from the original 3D-Fauna dataset and $0.4$ from our mouse dataset. We freeze the base shape bank while fine-tuning the instance deformation network, pose predictor, and texture field. Optimization is performed for 500k iterations with a batch size of 10, using Adam (learning rate $1\times 10^{-4}$ for the instance network, $1\times 10^{-3}$ for the base network, weight decay $0$). However, we observe that 3D-Fauna struggles to fit our mouse dataset, which covers the entire viewing sphere, leading to suboptimal performance in multi-view reconstruction.

Finally, we benchmark inference efficiency: MoReMouse runs at approximately 0.91 seconds per frame on an NVIDIA A100 GPU, consuming 7.9 GB of GPU memory during inference.

\subsection{Comparison on Original Images of ``markerless\_mouse\_1''}
In addition to the evaluations on the \textit{Synthetic Evaluation Data} constructed from the last 6{,}000 frames of the ``markerless\_mouse\_1'' sequence,  
we further evaluate all methods by directly reconstructing the same 6{,}000 frames from the original 6-view captures of ``markerless\_mouse\_1''.  
Qualitative comparisons are shown in Fig.~\ref{fig:large_compare_ori}, and quantitative results are summarized in Table~\ref{tab:origin_data_results}. 

\begin{table}[h]
\centering
\renewcommand{\arraystretch}{1.2}
\setlength{\tabcolsep}{4pt}
\begin{tabular}{lcccc}
\toprule
\textbf{Method} & \textbf{PSNR} $\uparrow$ & \textbf{SSIM} $\uparrow$ & \textbf{LPIPS} $\downarrow$ & \textbf{IoU} $\uparrow$ \\
\midrule
Ours & \cellcolor{cyan!25}17.987 & \cellcolor{cyan!25}0.9352 & \cellcolor{cyan!25}0.1001 & \cellcolor{cyan!25}0.6071 \\
Ours w/o embed & \cellcolor{cyan!10}17.929 & \cellcolor{cyan!10}0.9346 & \cellcolor{cyan!10}0.1010 & \cellcolor{cyan!10}0.6001 \\
TripoSR & 15.837 & 0.9202 & 0.1453 & 0.4474 \\
TripoSR-Tuned & \cellcolor{cyan!5}17.788 & \cellcolor{cyan!5}0.9333 & \cellcolor{cyan!5}0.1060 & \cellcolor{cyan!5}0.5993 \\
Triplane-GS & 15.703 & 0.9182 & 0.1398 & 0.4682 \\
Triplane-GS-Tuned & 17.549 & 0.9293 & 0.1117 & 0.5966 \\
LGM & 14.079 & 0.9051 & 0.1519 & 0.3230 \\
InstantMesh & 14.876 & 0.9127 & 0.1313 & 0.4080 \\
Fauna & 14.261 & 0.9086 & 0.1488 & 0.4030 \\
Fauna-Tuned & 14.054 & 0.9089 & 0.1481 & 0.2956 \\
TRELLIS & 13.199 & 0.9146 & 0.1304 & 0.1242 \\
\bottomrule
\end{tabular}
\caption{\textbf{Quantitative results on the original 6-view ``markerless\_mouse\_1'' dataset.}  
Cyan shading indicates top-3 performance.}
\label{tab:origin_data_results}
\end{table}

\subsection{Comparison on Real Data}
we evaluate all compared approaches on our captured real multi-view dataset. Qualitative comparisons are presented in Figure~\ref{fig:large_compare_real}, and quantitative results are summarized in Table~\ref{tab:real_data_results}. 
Our method consistently achieves the best performance across all three metrics, indicating superior reconstruction fidelity and perceptual quality.

\begin{table}[h]
\centering
\renewcommand{\arraystretch}{1.2}
\setlength{\tabcolsep}{4pt}
\begin{tabular}{lcccc}
\toprule
\textbf{Method} & \textbf{PSNR} $\uparrow$ & \textbf{SSIM} $\uparrow$ & \textbf{LPIPS} $\downarrow$ & \textbf{IoU} $\uparrow$ \\
\midrule
Ours & \cellcolor{cyan!25}18.422 & \cellcolor{cyan!25}0.9478 & \cellcolor{cyan!25}0.08674 & \cellcolor{cyan!25}0.6353 \\
Ours w/o embedding & \cellcolor{cyan!10}18.250 & \cellcolor{cyan!10}0.9469 & \cellcolor{cyan!10}0.08767 & \cellcolor{cyan!10}0.6291 \\
TripoSR & 11.518 & 0.8114 & 0.19672 & 0.4741 \\
TripoSR-Tuned & \cellcolor{cyan!5}18.162 & \cellcolor{cyan!5}0.9461 & \cellcolor{cyan!5}0.09354 & \cellcolor{cyan!5}0.6215 \\
Triplane-GS & 16.789 & 0.9298 & 0.11002 & 0.5743 \\
Triplane-GS-Tuned & 18.628 & 0.9469 & 0.09239 & 0.6099 \\
LGM & 15.215 & 0.9197 & 0.12838 & 0.4138 \\
InstantMesh & 15.631 & 0.9175 & 0.11339 & 0.5466 \\
3D Fauna & 13.284 & 0.8837 & 0.12890 & 0.4467 \\
3D Fauna-Tuned & 13.726 & 0.8977 & 0.12038 & 0.4254 \\
TRELLIS & 15.087 & 0.9279 & 0.10968 & 0.3095 \\
\bottomrule
\end{tabular}
\caption{\textbf{Quantitative results on the real-world multi-view dataset.}
Cyan shading indicates top-3 performance.}
\label{tab:real_data_results}
\end{table}

\section{Failure cases}

\begin{figure}[h]
  \centering
  \includegraphics[width=1.0\linewidth]{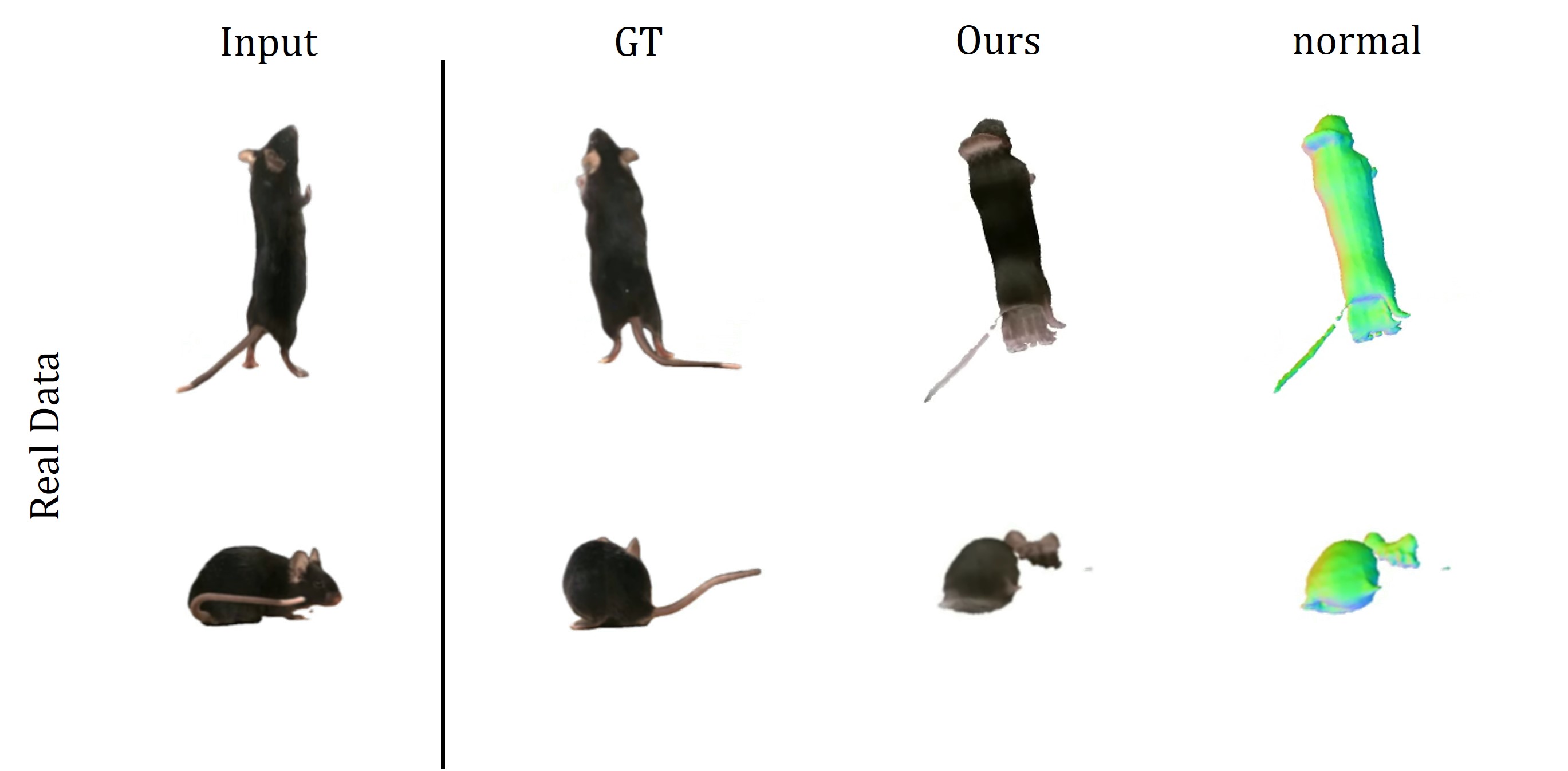}
  \caption{Representative failure cases. Top: failure due to lack of standing mouse data in training, leading to unrealistic reconstructions. Bottom: failure due to tail-induced occlusion from shaved fur, which interferes with head feature extraction.}
  \label{fig:failure_case}
\end{figure}

Figure~\ref{fig:failure_case} showcases typical failure cases on real-world data. The first row highlights the model’s inability to generalize to rare upright postures, due to the absence of such poses in the training set. This leads to implausible geometry and emphasizes the need for more pose-diverse avatar data.

In the second row, reconstruction fails when the tail occludes the head—especially under shaved-fur conditions—causing the model to misinterpret head structure. While our preprocessing removes background, we do not explicitly account for occlusions from scene objects or self-occlusion, which can degrade reconstruction fidelity.

Additionally, we observe that small regions such as feet and snouts are more prone to distortion, likely due to limited resolution in those areas. Future work may explore resolution-aware refinement or occlusion-robust modeling to address these limitations.

\section{More Details about MoReMouse Training}
Our model is trained in two stages: the NeRF-based volumetric rendering stage is trained for $60$ epochs to ensure proper density field convergence, followed by the DMTet-based explicit geometry stage trained for $100$ epochs to refine surface details. 
The overall optimization objective combines geometric, photometric, and perceptual terms, and is defined as:
\begin{equation} 
\begin{aligned}
\mathcal{L} &= \lambda_{\text{MSE}} \mathcal{L}_{\text{MSE}} + \lambda_{\text{depth}} \mathcal{L}_{\text{depth}} + \lambda_{\text{smoothL1}} \mathcal{L}_{\text{smoothL1}} \\
&\quad + \lambda_{\text{mask}} \mathcal{L}_{\text{mask}} + \lambda_{\text{LPIPS}} \mathcal{L}_{\text{LPIPS}}.
\end{aligned}
\end{equation}
Here, \( \mathcal{L}_{\text{MSE}} \) enforces pixel-wise supervision for both RGB values and feature embeddings, while \( \mathcal{L}_{\text{depth}} \) ensures depth consistency within the valid object region. 
The smooth L1 loss \( \mathcal{L}_{\text{smoothL1}} \) penalizes large RGB discrepancies, and \( \mathcal{L}_{\text{mask}} \) applies binary cross-entropy supervision to the predicted opacity maps. 
The perceptual loss \( \mathcal{L}_{\text{LPIPS}} \) measures high-level perceptual similarity in a deep feature space. 
The hyperparameters \( \lambda \) control the relative importance of each term.

We summarize the architectural and training hyperparameters of the MoReMouse model in Table~\ref{tab:model_details}, covering key components such as the image encoder, triplane representation modules, multi-head MLP, NeRF and DMTet renderers, and training configurations.

\begin{table*}[h]
\centering
\renewcommand{\arraystretch}{1.2}
\begin{tabular}{lll}
\toprule
\textbf{Module} & \textbf{Parameter} & \textbf{Value} \\
\midrule
\multirow{2}{*}{Image Tokenizer}
& Type & dinov2-base \\
& Image resolution & $378 \times 378$ \\
& Feature channels & $768$ \\
\midrule
\multirow{2}{*}{Triplane Tokenizer} 
& Plane size & $64 \times 64$ \\
& Channels & $512$ \\
\midrule
\multirow{5}{*}{Backbone} 
& Input channels & $512$ \\
& Attention layers & $12$ \\
& Attention heads & $16$ \\
& Attention head dimension & $64$ \\
& Cross attention dimension & $768$ \\
\midrule
\multirow{3}{*}{Triplane Upsampler} 
& Input channels & $512$ \\
& Output channels & $80$ \\
& Output shape & $3 \times 80 \times 128 \times 128$ \\
\midrule
\multirow{4}{*}{MultiHeadMLP} 
& Neurons & $64$ \\
& Shared hidden layers & $10$ \\
& Hidden layers for density head & $1$ \\
& Hidden layers for feature/deformation heads & $1$ \\
\midrule
\multirow{4}{*}{Renderer-NeRF} 
& Samples per ray & $128$ \\
& Radius & $0.87$ \\
& Density activation & \texttt{trunc\_exp} \\
& Density bias & $-1.0$ \\
\midrule
\multirow{2}{*}{Renderer-DMTet} 
& SDF bias & $-4.0$ \\
& Isosurface resolution & $256^3$ \\
\midrule
\multirow{10}{*}{Training} 
& Learning rate & $1\text{e}{-5}$ \\
& Optimizer & AdamW \\
& LR scheduler & CosineAnnealingLR \\
& Warm-up steps & $3000$ \\
& $\lambda_{\text{mse}}$ & $1.0$ \\
& $\lambda_{\text{LPIPS}}$ & $1.0$ \\
& $\lambda_{\text{mask}}$ & $0.3$ \\
& $\lambda_{\text{smooth}}$ & $0.2$ \\
& $\lambda_{\text{depth}}$ & $0.2$ \\
\bottomrule
\end{tabular}
\caption{MoReMouse Architecture and Training Hyperparameters}
\label{tab:model_details}
\end{table*}

\end{document}